\documentclass[preprint,12pt,authoryear]{elsarticle}

\usepackage{xcolor}
\usepackage{amsmath}
\usepackage{algorithm}
\usepackage{algpseudocode}
\newcommand{\ccc}[1]{\textcolor{magenta}{#1}}

\newcommand{\omitme}[1]{}

\usepackage[export]{adjustbox}
\usepackage{booktabs}
\usepackage{float}
\usepackage{verbatim}
\usepackage{gensymb}
\usepackage{subcaption}
\usepackage{siunitx}
\usepackage[hyphens]{url}
\usepackage[hidelinks]{hyperref}
\hypersetup{breaklinks=true}
\usepackage{amssymb}
\usepackage{setspace}
\usepackage[a4paper,margin=2.5cm]{geometry}

\journal{Remote Sensing of Environment}

\begin{document}

\begin{frontmatter}

\title{Very high resolution canopy height maps from RGB imagery using self-supervised  vision transformer and convolutional decoder trained on Aerial Lidar}

\author{Jamie Tolan$^{1}$, Hung-I Yang$^{1}$, Benjamin Nosarzewski$^{1}$, Guillaume Couairon$^{2}$ , Huy V. Vo$^{2}$ , John Brandt$^{3}$, Justine Spore$^{3}$, Sayantan Majumdar$^{4}$, Daniel Haziza$^{2}$, Janaki Vamaraju$^{1}$, Theo Moutakanni$^{2}$, Piotr Bojanowski$^{2}$, Tracy Johns$^{1}$, Brian White$^{1}$, Tobias Tiecke$^{1}$, Camille Couprie$^{2}$}

\address{%
$^{1}$  Meta;  
$^{2}$  Fundamental AI Research (FAIR), Meta; 
$^{3}$  World Resources Institute;
$^{4}$  Desert Research Institute
}

\begin{abstract}
Vegetation structure mapping is critical for understanding the global carbon cycle and monitoring nature-based approaches to climate adaptation and mitigation. Repeated measurements of these data allow for the observation of deforestation or degradation of existing forests, natural forest regeneration, and the implementation of sustainable agricultural practices like agroforestry. 
Assessments of tree canopy height and crown projected area at a high spatial resolution are also important for monitoring carbon fluxes and assessing tree-based land uses, since forest structures can be highly spatially heterogeneous, especially in agroforestry systems. Very high resolution satellite imagery (less than one meter (1m) Ground Sample Distance) makes it possible to extract information at the tree level while allowing monitoring at a very large scale. This paper presents the first high-resolution canopy height map concurrently produced for multiple sub-national jurisdictions. 
Specifically, we produce very high resolution canopy height maps for the states of California and S\~{a}o Paulo, a significant improvement in resolution over the ten meter (10m) resolution of previous Sentinel / GEDI based worldwide maps of canopy height. 
The maps are generated by the extraction of features from a self-supervised model trained on Maxar imagery from 2017 to 2020, and the training of a dense prediction decoder against aerial lidar maps. We also introduce a post-processing step using a convolutional network trained on GEDI observations.
We evaluate the proposed maps with set-aside validation lidar data as well as by comparing with other remotely sensed maps and field-collected data, and find our model produces an average Mean Absolute Error (MAE) of 2.8 meters and Mean Error (ME) of 0.6 meters. 
\end{abstract}

\end{frontmatter}

\section{Introduction}

Spatially explicit maps of forest vegetation structure, such as tree canopy height and crown projected area, are powerful tools for assessing forest degradation, forest and landscape restoration (FLR), and estimating above-ground woody biomass for carbon emission and sequestration modeling. Existing assessments of the climate implications of woody vegetation flux, including FLR, deforestation, and natural regrowth, often rely on remotely sensed dynamic vegetation models of deforestation and regrowth \citep{Friedlingstein-2019}. Such wall-to-wall data on tree height and canopy structure are used to estimate aboveground woody biomass. However, land-use patterns operate on more granular spatio-temporal scales than those captured by global carbon models, which typically have coarse spatio-temporal resolution. This contributes to the large uncertainty in existing nation-wide and global accounting of carbon stored in forests \citep{Popkin2015Hunt, DUNCANSON2020111779, Yanai_2020}. For instance, \cite{Cook-Patton2020} produce a global 1-km scale map of potential above-ground carbon accumulation rates by developing machine learning models based on more than 13,000 locations derived from literature. \cite{Cook-Patton2020} find significant variability in predicted carbon accumulation rates compared to defaults from the International Panel on Climate Change (IPCC) at the ecozone scale. In the African tropical montane forests, \cite{Cuni-Sanchez2021} model forest carbon density based on 72,336 measurements of height and tree diameter, identifying two-thirds higher carbon stocks than the respective IPCC default values. 

The uncertainty of biomass modelling also affects the uncertainty of the carbon implications of deforestation and regrowth. Tree-based FLR, including agroforestry, reforestation, natural regeneration, and enrichment planting, is considered to be a cost-effective natural climate solution for adaptation and mitigation. However, evaluating the effectiveness of FLR interventions at a large scale is difficult due to its highly distributed nature, typically being practiced on individual land parcels by respective land owners \citep{reytar2020}. While carbon reporting frameworks exist for FLR, for example through verified carbon markets, such data are highly project-specific owing to their reliance on intensive manual field measurements. Utilizing remotely sensed data to assess vegetation structure on areas with FLR interventions such as intercropped agroforestry or natural regeneration is difficult due to the presence of multiple species, multiple canopy strata, and trees of different ages \citep{viani2018, Vallauri2005, Camarretta2020}. For instance, \cite{fikrey2022} found that 70\% of the shade trees in an agroforestry system in Ethiopia were below 3 meters in height, while 3\% were above 12 meters in height, with more than a two-order of magnitude range of per-tree carbon stocks depending on tree size.

Critical to reducing uncertainty in woody carbon models are measurements of forest height and biomass to improve assessments of the spatial variability of carbon removal rates across forest landscapes that have heterogeneous structure \citep{Harris2021}. Tree height is especially critical to accurately assessing carbon removal rates, as growth rate increases continuously with size \citep{Stephenson2014}. Recent earth observation missions from NASA, namely GEDI and ICESat-2, provide repeated vegetation canopy height maps for the first time. Global Ecosystem Dynamics Investigation (GEDI) collects canopy height and relative height at a 25 meter resolution \citep{gedi}. ICESat-2 collects canopy height and relative height at a $13 \times 100$ meter native footprint \citep{MARKUS2017260}. Recently, multi-sensor fusion has demonstrated potential to improve aboveground biomass mapping \citep{SILVA2021112234}. To generate wall-to-wall maps of canopy height, researchers commonly combine active optical LiDAR data from ICESat-2 or GEDI with optical imagery from Sentinel-2 \citep{Lang2022High, Schwartz2021High} or Landsat satellites \citep{Schwartz2021High, LI2020102163}.

A number of recent studies have utilized spaceborne lidar data from GEDI and ICESat-2 to produce canopy height maps in combination with multispectral optical imagery. Among them, \cite{Potapov2021Mapping} combined GEDI RH95 ($95^{\mbox{th}}$ percentile of Relative Height) data with Landsat data to establish a global map at 30 meter resolution, using a bagged regression tree ensemble algorithm. More recently, \cite{Lang2022High} produced a global canopy height map at a 10-meter resolution, applying an ensemble of convolutional neural network (CNN) models to Sentinel-2 imagery to predict the GEDI RH98 footprint. Other works have produced regional 10-meter CHMs utilizing Sentinel-2 and aerial lidar \citep{rs13122392, fayad2023vision}.

Aerial lidar data has also demonstrated utility as training data for high resolution ($<$ 5 m) and very high resolution ($<$ 1 m) canopy height maps.  At a national scale, \cite{Csillik2019} generated biomass maps in Peru by applying gradient boosted regression trees between 3.7 meter Planet Dove imagery and airborne lidar, with low uncertainty in dense forests but large amounts of uncertainty in transitional landscapes and areas that are hotspots of land use change. Recently, \cite{Siyu2023overlooked} computed a canopy height map (CHM) map of Europe using 3 meter Planet imagery, training two UNets to predict tree extent and CHM using lidar observations and previous CHM predictions from the literature. Utilizing aerial optical imagery, \cite{wagner2023submeter} generated a submeter CHM of California, USA by training a regression U-Net CNN on 60-cm imagery from the USDA-NAIP program and aerial lidar. 

The estimation of canopy height from high resolution optical imagery shares similarities with the computer vision task of monocular depth estimation. Vision transformers, which are a deep learning approach to encoding low-dimensional input into a high dimensional feature space, have established new frontiers in depth estimation compared to convolutional neural networks \citep{Ranftl2021DPT}. While depth estimation models benefit significantly from large receptive fields \citep{LI2018328, fu2018compromise, Miangoleh_2021_CVPR}, \cite{https://doi.org/10.48550/arxiv.1701.04128} demonstrate that the effective receptive fields of CNN models have Gaussian distributions, limiting the ability for CNNs to model long-range spatial dependencies. In contrast to convolutional neural networks (CNNs), which subsequently apply local convolutional operations to enable the modelling of increasingly long-range spatial dependencies, transformers utilize self-attention modules to enable the modeling of global spatial dependencies across the entire image input \citep{https://doi.org/10.48550/arxiv.2010.11929}. 

For dense prediction tasks on high resolution imagery where the context can be sparse, such as ground information in the case of near closed canopies, the ability of transformers to model global information is promising. Among the applications to aerial imagery, the work of \cite{rs13183585} uses a Swin transformer to classify high-resolution land cover. Finding that a baseline transformer model struggled with edge detection, \cite{rs13183585} utilized a self-supervised edge extraction and enhancement method to improve definition of class edges. \cite{9810316} utilize the vision transformer architecture as a feature encoder, and apply a feature pyramid decoder to the resulting multi-scale feature maps. \cite{drones7020093} segment individual date palm trees by applying vision transformers to 5- to 30-cm drone-based imagery, finding that the Segformer architecture improves generalizability to different resolution imagery when compared to CNN-based models. More recently, also leveraging vision transformers, \cite{reed2022scale} scale the Masked Auto-Encoder approach of \cite{he2022masked} and apply it to building segmentation.

A major challenge of applying high resolution, airborne lidar data to the generation of wall-to-wall canopy height maps is the relative scarcity of airborne lidar data to the scientific community. Such scarcity can negatively impact the generalizability of models to unseen geographies, especially data-poor regions where little to no airborne lidar exists \citep{rs15051407}. Given this context of low annotation, Self-Supervised Learning (SSL) is a promising tool to shape more robust features than traditional deep approaches. In particular, the SSL DINOv2 approach of \cite{Oquab2023SSL} recently led to state-of-the-art performances in several computer vision tasks such as image classification, depth prediction, and segmentation. In the context of satellite image analysis, self-supervised learning has been shown to improve the generalizability of building segmentation models in Africa \citep{Sirko2021Continental}. To mitigate the reliance of vision transformers on self-supervised learning, \cite{fayad2023vision} utilized knowledge distillation with a U-Net CNN teacher model to generate 10-meter CHM of Ghana using Sentinel-1, Sentinel-2, and aerial lidar. 

Understanding the importance of highly spatially explicit vegetation structure mapping to both large-scale carbon modeling and project-specific avoided deforestation and restoration monitoring, the objective of this study is to produce high resolution canopy height maps that are able to scale and generalize to large geographic regions. Our method consists of an image encoder-decoder model, where low spectral dimensional input images are transformed to a high dimensional encoding and subsequently decoded to predict per-pixel canopy height. We employ DINOv2 self-supervised learning to generate universal and generalizable encodings from the input imagery \citep{Oquab2023SSL}, and train a dense vision transformer decoder \citep{Ranftl2021DPT} to generate canopy height predictions based on aerial lidar data from sites across the USA. To correct a potential bias coming from a geographically limited source of supervision, we finally refine the maps using a convolutional network trained on spaceborne lidar data. We present canopy height maps for the states of S\~{a}o Paulo, Brazil, and California, USA, and provide qualitative and quantitative error analyses of height estimation and the decomposition of height estimates into tree segmentation maps.

\section{Data}
\label{sec:datasets}

\subsection{Experimental Design}
\label{sec:expdesign}
This paper presents canopy height maps for S\~{a}o Paulo State, Brazil, and California State, USA. These geographies were chosen due to their prevalence of timber production, presence of old growth forests, mountainous terrains, and high degree of tree biodiversity \citep{Maioli_Belharte_StukerKropf_Callado_2020, Luyssaert2008, Ribeiro2011}. The dataset was generated with a machine learning model utilizing a transformer encoder and convolutional decoder trained with an input composite of approximately 0.59 meter GSD Maxar imagery spanning the years 2018 to 2020 and output labels from 1 meter GSD aerial lidar. Our data and methods sections are structured as follows. First, we describe the satellite and aerial lidar data used for model training and map generation. Next, we describe the model training specifics, including self supervised learning and the methods for combining models trained on aerial lidar with models trained on GEDI observations, and the baseline models selected and ablation studies performed. Finally, we present our approach for qualitative and quantitative evaluation of height accuracy and tree segmentation, and discuss the generalization of our model.

\subsection{Satellite image data description}
\label{sec:satdata}

Maxar Vivid2 mosaic imagery\footnote{\url{https://resources.maxar.com/data-sheets/imagery-basemaps-data-sheet}} served as input imagery for model training and inference. This dataset provides global coverage by mosaicing together imagery from multiple instruments (WorldView-2 (WV 2), WorldView-3 (WV 3), Quickbird II) and observation dates. By starting with this mosaiced imagery, we leveraged the extensive data selection pipeline from Maxar, resulting in imagery that had less than $2\%$ percent cloud cover, a global revisit rate predominately (more than $75\%$) below 36 months (imagery dates from 2017 to 2020 are utilized in this dataset), view angles of less than 30 degrees off nadir, and sun angle of less than 60 degrees from zenith. This imagery consisted of three spectral bands: Red, Green, and Blue (RGB), with approximately a 0.5 meter GSD. The imagery was processed in the Web Mercator projection (EPSG:3857) and stored with the Bing tiling scheme\footnote{\url{https://learn.microsoft.com/en-us/bingmaps/articles/bing-maps-tile-system}}. Given the high resolution of the original geotiffs, Bing zoom 15 level tiles, with $2048 \times 2048$ pixels per tile  were used, giving a pixel size of 0.597 meters GSD at the equator. 

\subsection{Satellite image data preparation} 

\subsubsection{Image preparation} 
\label{sec:data_prep}

For easier training and validation of computer vision models, we extracted small regions from the input satellite imagery. Centered around a given location, a box of fixed ground distance was selected, using a local tangent plane coordinate system. Due to the Web Mercator projection of the image tiles, the extracted images at each position had varying dimensions according to their latitude, which were re-sampled to a fixed number of pixels. We chose a box side length of 152.7 meters, which, when re-sampled to 256$\times$256 pixel images, provided ``thumbnail" images that matches the lowest resolution (0.597m) of the input imagery described in Section \ref{sec:satdata}. Using these thumbnail images both for training and inference ensured that the dataset had constant number of pixels and that the pixel size was the same for all latitudes, preventing potential biases with latitude which may be introduced by variation in pixel size.

\subsubsection{Dataset for Self-Supervised Learning}
\label{sec:ssl_data}
For training the self-supervised encoder, we randomly sampled 18 million $256\times256$ pixel satellite thumbnail images. No labels were used for the SSL stage. 

\subsubsection{Validation segmentation dataset}
\label{sec:seg_data}
We also manually annotated a random selection of 9000 Maxar thumbnail images for segmentation testing. A binary tree / no tree label was applied by human annotators. Pixels estimated to have a canopy height above one meter (1m) tall and with a canopy diameter of more than three meters (3m) were labeled as tree.

\subsection{Supervised dataset}

\label{sec:als_data}

We gathered approximately 5800 canopy height maps (CHM), selected from the \cite{NeonData}. Each CHM typically consisted of 1km $\times$ 1km geotiffs, with a pixel size of one meter (1m) GSD, in local UTM coordinates.  
We selected the sites used by \cite{Weinstein2021Benchmark} and additionally manually filtered for sites that have CHM imagery that was well registered and mostly free from mosaicing artifacts. Additionally, we selected sites with imagery acquired less than two years from the observation date in the associated Maxar satellite imagery. A complete list of NEON sites used for training and validation is contained in \ref{appendix:neonsites}.

The CHM geotiffs were reprojected to a local tangent plane coordinate system and resized to match the resolution of Maxar images. For each ALS CHM, a corresponding RGB satellite image was linked, and these pairs of imagery served as the training data for our decoder model. The 5800 images in the NEON ALS dataset were split in sets of $80\%$ training images, $10\%$ calibration and $10\%$ set-aside validation images. During the training, validation and testing phases, we sampled $256\times256$ random crops from the RGB - ALS image pairs. Model training was conducted over epochs sampled from the training dataset. At the completion of each epoch, metrics were computed from a 10\% calibration dataset to calibrate the hyperparameters of the model training process. The calibration dataset was drawn from the same set of sites as the training datasets, but from separate 1km $\times$ 1km geotiffs to ensure non overlapping pixels.

We constructed a set-aside validation dataset from a subset of sites in our NEON dataset, which we call ``NEON test". None of the sites used in the validation dataset were contained in the training or calibration dataset. A list of NEON sites in the validation set appears in \ref{appendix:neonsites}. We also prepared two validation datasets from other publicly available ALS Lidar datasets, outside of the NEON collection. These datasets covered different geographic locations and ecosystems: ``CA-Brande" \citep{CADataset} covered a coastal ecosystem in CA, and ``S\~{a}o Paulo" \citep{SaoPauloDataset} covered a region in the Brazilian S\~{a}o Paulo State. 
See Figure \ref{fig:dataset_distributions} for a visual breakdown of the Neon dataset splits.

Where these datasets were available as CHMs, we directly used the supplied CHMs. However, for the S\~{a}o Paulo datasets, which only contained point cloud datasets, we processed CHMs following the pit-free algorithm \citep{Khosravipour2014}. The pit-free algorithm was also adopted by the NEON team for generating their CHM product, and we found that different input parameters to the pit-free algorithm had negligible impact on the CHM output.

\subsection{Data Augmentation}

The $256\times256$ pixel image thumbnail images of RGB and CHM imagery were augmented at training time, with random 90 degree rotations, brightness, and contrast jittering. We found that these augmentations improved model prediction stability across the various Maxar observations in the input dataset. 

\section{Model and data generation methods}
\label{sec:methods}

Our goal was to create a model that produces high resolution canopy height maps and generalizes across large geographic scales. To accomplish that goal, we leveraged the relative strengths of two types of lidar data. Aerial lidar provided high resolution canopy height estimation, but lacks global spatial coverage. In comparison, GEDI has nearly global coverage of transects, but its beam width of approximately 25 meters did not allow for the identification of individual trees.

\begin{figure}[htb]
  \centering 
  \includegraphics[width=1.0\textwidth]{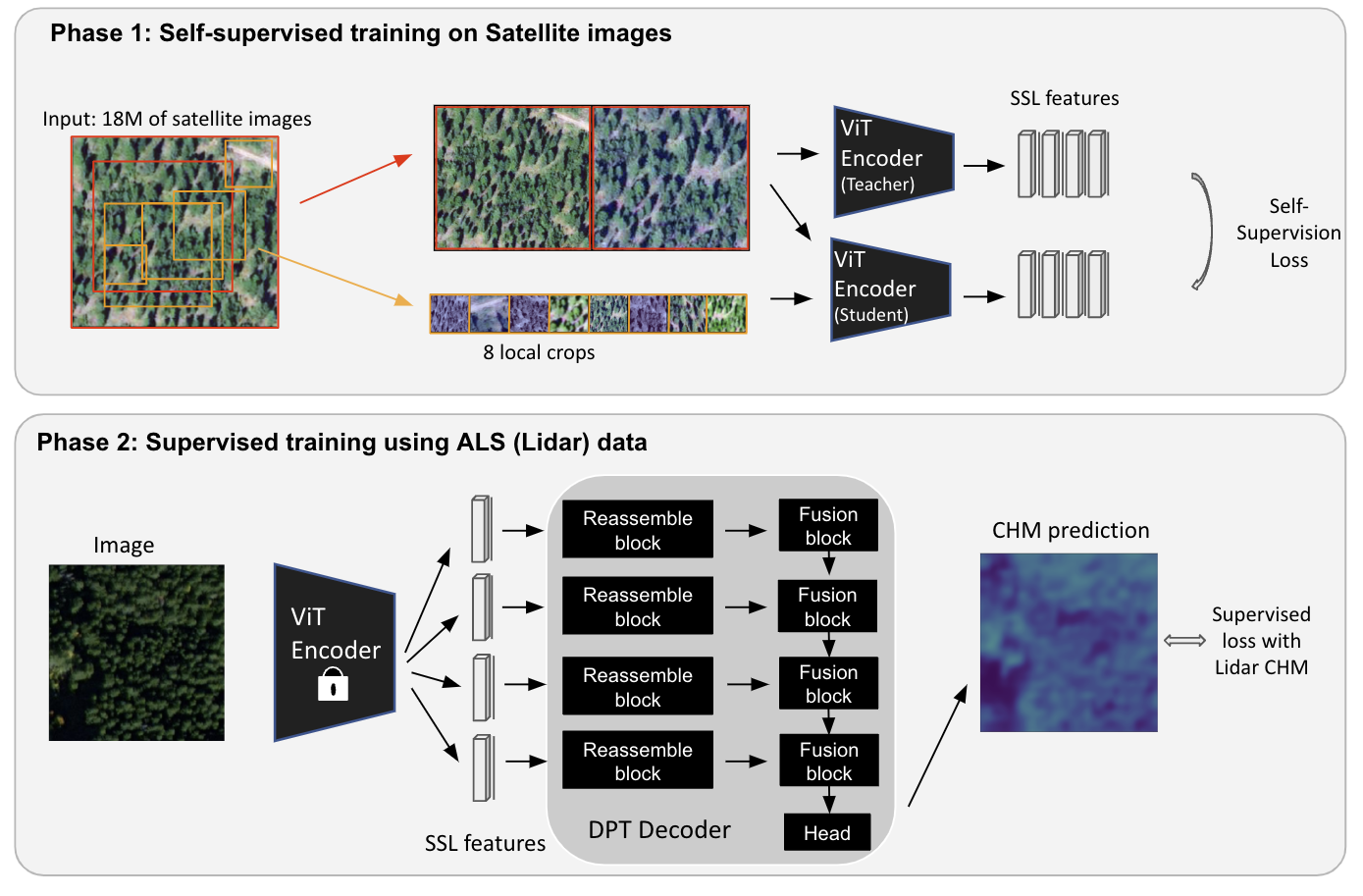}
  \caption{Overview of our approach for generating ALS-based CHMs. During the first stage, we employed the self-supervised learning approach \cite{Oquab2023SSL} on 18 million $256\times256$ satellite images leading to a set of four spatial feature maps, and four feature vectors, extracted at different layers of the Vision Transformer model (ViT). In the second phase, we trained a convolutional DPT decoder to predict CHMs.} 
  \label{fig:model}
\end{figure}

After self-supervised pre-training on satellite images globally, our high-resolution ALS CHM prediction model was trained on images from the NEON dataset, as detailed in Section \ref{sec:alsmodel} and Figure \ref{fig:model}. As the Neon dataset only has a spatial coverage from sites only within the United States, we expect this ALS CHM model to perform well on ecosystems similar to the training set. To improve generalization of other ecosystems and locations, a low resolution CHM model was independently trained on global GEDI data (Section \ref{sec:gedimodel}). The GEDI model was used to compute a rescaling factor map (Section \ref{sec:gedicorr}), which adjusted the predictions made by the ALS CHM model.

\subsection{Self Supervised Learning}
\label{sec:ssl}

Following the recent success of self-supervised learning on dense prediction tasks from \cite{Oquab2023SSL}, we employed a self-supervised learning step on 18 million globally distributed, randomly sampled $256 \times 256$ pixel Maxar satellite images to obtain an image encoder delivering features specialized to vegetation images. In the training phase, different views of the image were fed to two versions of the encoder: a teacher model receiving global crops, and a student model receiving local and global views where part of the crops were masked (replaced by zero values). We employ a huge ViT architecture, where the inputs are decomposed into $16\times 16$ patches. The two networks were trained jointly to output similar feature representations. The procedure is illustrated in the Phase 1 in Figure~\ref{fig:model}. In a second phase described in Section \ref{sec:alsmodel}, we freeze the SSL encoder layers using the weights of the teacher model and train the decoder with ALS data to generate high-resolution canopy height maps.

\subsection{High resolution canopy height estimation using ALS}
\label{sec:alsmodel}

We used the reference dataset described in Section \ref{sec:als_data}, prepared following the methods described in Section \ref{sec:data_prep}. The output of the ALS model was a raster of predicted canopy heights at the same resolution as the input imagery. For training the supervised decoder, we used the ALS CHM data described in Section \ref{sec:als_data} to create a connection between the SSL features and the full resolution canopy height image. In this second phase, we trained the decoder introduced in Dense Prediction Transformer (DPT) \citep{Ranftl2021DPT} on top of the obtained features. This approach is described in Figure \ref{fig:model}, phase 2. The DPT paper describes a full model composed of a transformer encoder extracting features at different layers. In the decoder, each output was reassembled and all outputs were fused. In our second phase of ALS training, we replaced the transformer of DPT by our own SSL encoder, and trained the DPT decoder part only, from scratch. Our best results were obtained by freezing all layers from the SSL encoder. We employed a one cycle learning rate schedule with a linear warmup in the encoder training stage and a ``Sigloss" loss function. Further architecture and training details are provided in \ref{App:training_details}.

\paragraph{Sigloss function}

We take advantage from the similarity of canopy height mapping to the task of depth estimation and borrow the loss from \cite{eigen2014depth}. Given a true canopy height map $c$ and our prediction $\hat{c}$, the Sigloss is given by    

\begin{equation}
\mathcal{L} = \alpha \sqrt{\frac{1}{T}\sum_i \delta_i^2 - \frac{\lambda}{T^2}(\sum_i \delta_i)^2},
\end{equation}
where $\delta_i = \log(\hat{c_i}) - \log(c_i)$, and $T$ is the number of pixels with valid ground truth values. As in previous works, we fix $\lambda=0.85$ and $\alpha = 10$. 

\paragraph{Classification output}

To avoid a bias toward small predicted values, we implemented a classification step first, combined with the Sigloss defined above. The strategy is described by \cite{bhat2021adabins} as the uniform strategy. Specifically, we modified the output of our decoder to return, instead of one scalar per pixel, a range of $B$ bins. After a normalization on the predictions, we computed the scalar product between the obtained histogram of predicted bins and a linear vector ranging [0,$B$], with $B$ set to 256.

\subsection{Large scale canopy height estimation using GEDI prediction model}
\label{sec:gedimodel}

\begin{figure}[htb]
  \centering 
  \includegraphics[width=1.0\textwidth]{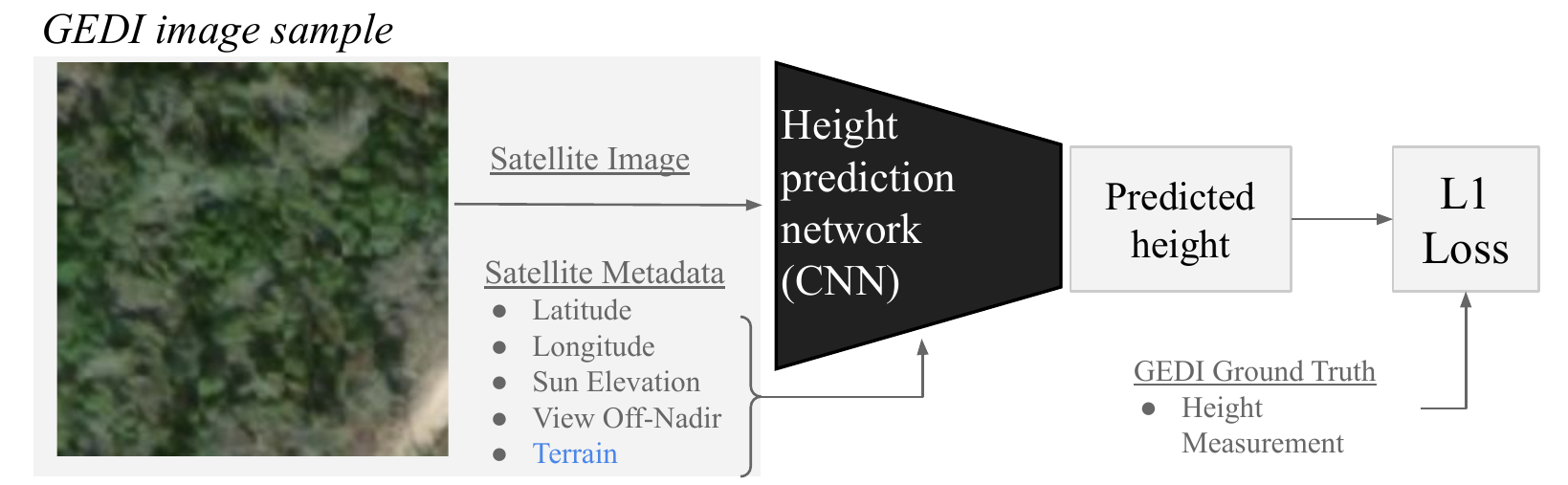}
  \caption{Overview of our methodology to generate predicted RH95 values using GEDI measurements across the globe. Terrain is used only during the training and set to zero during inference.}
  \label{fig:train}
\end{figure}

To mitigate the effect of the limited geographic distribution of available ALS data, we employed a second regression network trained on GEDI data to rescale the ALS network outputs. The GEDI prediction model was a simple convolutional network, containing five convolutional layers, followed by five fully connected layers. The inputs to the model were $128\times128$ pixel Maxar images containing three RGB bands, in topocentric coordinates, processed as described in Section \ref{sec:data_prep}. The ground truth data consisted of 13 million GEDI measurements, which were randomly sampled from the full GEDI dataset described in \ref{sec:gedi_data}. 
We trained the GEDI model to output a single scalar value for a 128$\times$128 pixel image patch, with a L1 loss on a regression task against the RH95 value from the GEDI instrument. The training details are specified in \ref{sec:gedi_training}.

\subsection{Combining ALS and GEDI model outputs}
\label{sec:gedicorr}

\begin{figure}[htb]
  \centering 
  \includegraphics[width=1.0\textwidth]{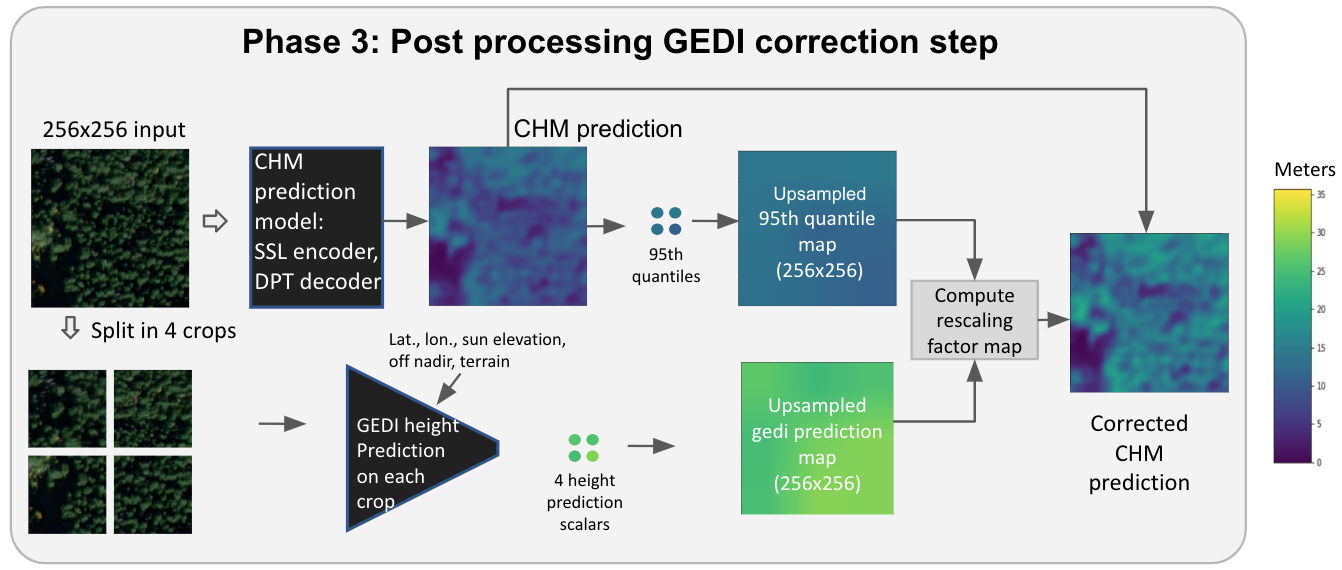}
  \caption{Post processing step using GEDI predictions during inference. We used the GEDI model to correct our CHM predictions, by computing a dense scaling factor, and multiply it pointwise with the CHM prediction map.}
  \label{fig:correction}
\end{figure}

In this section, we describe the process of connecting our GEDI model outputs (Section \ref{sec:gedimodel}) with ALS model outputs (\ref{sec:alsmodel}). Conceptually, the ALS model output provides high resolution canopy estimates but lacks the global context to correctly estimate the absolute height of vegetation in different ecosystems. Conversely, the GEDI model is trained on a global dataset and contains position and metadata inputs (Figure \ref{fig:train}). A schematic of the process is shown in Figure~\ref{fig:correction}.

\paragraph{Correlation between different lidar sources} The first step in making the GEDI/ALS connection is understanding the relationship between the two sets of lidar data: ALS CHM (Section \ref{sec:als_data}) and GEDI lidar (Section \ref{sec:gedi_data}). These two datasets make measurements of fundamentally different properties of canopy structure. GEDI measures the relative height of canopy based on the full waveform measurement of the return energy from 25 m diameter beam footprints while aerial lidar constructs higher resolution point clouds. To connect these two, we ran simulations with the GEDI simulator from \cite{Hancock2019} on the NEON ALS point clouds. We found that there was a strong correlation ($R^2=0.88$) between the 95th percentile of ALS canopy height maps and the simulated GEDI RH95 (see \ref{sec: connection between ALS P95 and GEDI RH95}). 

\paragraph{GEDI based correction of ALS trained maps}
We used this correlation to scale the ALS model canopy height maps by computing a scalar multiplier that match percentiles of the CHM map with the GEDI model predicted value for GEDI RH95. This process works as follows:

Given an input RGB image, $x$, we combined the outputs of the ALS and GEDI models by computing a dense correction factor $\gamma(x)$, so that the novel prediction, $C'(x)$ was related to the ALS model CHM, $C(x)$: 
\begin{equation}
C'(x) = \gamma(x) \odot C(x)
\end{equation}
where
\begin{equation}
\label{eqn:gamma}
\gamma(x) = \frac{1 + s_{\sigma}(G(x))}{1 + (s_{\sigma}((Q(x)_{95}))}.
\end{equation}
Here $G(x)$ is the output CHM of our GEDI model and $Q(x)_{95}$ is a per block upsampled $95^{\mbox{th}}$ percentile of the ALS model CHM in meters, computed over the exact same $128\times128$ pixel input regions as the input to the GEDI model in $G(x)$. More specifically, each input image was divided in four crops, each one independently fed to the height prediction model, leading to four scalars, that were concatenated and upsampled. From the predicted CHM map by our ALS model, we computed four percentiles from the same crops, concatenated and upsampled in the same way.

We used the ratio in Equation~\ref{eqn:gamma} rather than $G(x)/Q(x)_{95}$ to down-weight noisy model estimates near zero canopy height. Since $G(x)$ and $Q(x)_{95}$ are lower resolution than $C(x)$, the correction factor map was upsampled to match the resolution of the ALS CHM, $C(x)$. The ALS and GEDI maps were smoothed with a 20 pixel sigma Gaussian kernel $s_{\sigma}$ to prevent sharp transitions, and the correction factor was clipped between 0.5 and 2 to avoid drastic rescaling. 

\subsection{Baselines}
\subsubsection{ResUNet-based approach}
\label{sec:resunet}

We utilized a ResUNet-18 architecture for our baseline \citep{DBLP:journals/corr/abs-1711-10684}, which is an encoder-decoder architecture predicting a $N \times N$ canopy height map from a $3 \times N \times N$ RGB image, with $N=256$. The baseline model was trained with the sigloss between the predicted and ground truth CHMs. We also experimented with a classification output, however we did not obtain improvements from this approach.

\subsubsection{Supervised Transformer-based approach}

To assess the benefit of the self supervised training phase on Satellite data, we consider a baseline given the state-of-the-art vision SWAG encoder \citep{singh2022revisiting}. We used the large version of this Vision Transformer (ViT) that was trained to perform hashtag prediction from Instagram images. At the time of writing this manuscript, this model was in the top ten models with highest accuracy on ImageNet, CUB, Places, and iNaturalist datasets, providing a warranty of feature quality. This model contains the same number of parameters as our SSL encoder, allowing a fair comparison in terms of model size.

\subsection{Data validation}
\label{sec:validation}
We evaluated the model performance against a variety of metrics, which we divided into two broad classes: (1.)  Metrics which primarily evaluated the accuracy of canopy height maps, which we call canopy height metrics (Section~\ref{sec:chm_metrics}), and (2.) Metrics which primarily evaluated the accuracy of image segmentation into tree or no tree pixels, which we call segmentation metrics (Section~\ref{sec:seg_metrics}). The set-aside validation dataset of ALS canopy height maps described in Section~\ref{sec:datasets} served as the primary dataset for all types of metrics. For the segmentation metrics, we also evaluated the model predictions against a dataset of human-annotated labels independently labeled by photo-interpretation of Maxar imagery (Section ~\ref{sec:man_annot}).

\section{Results}

\begin{figure}[ht]
  \centering 
  \includegraphics[width=1.0\textwidth]{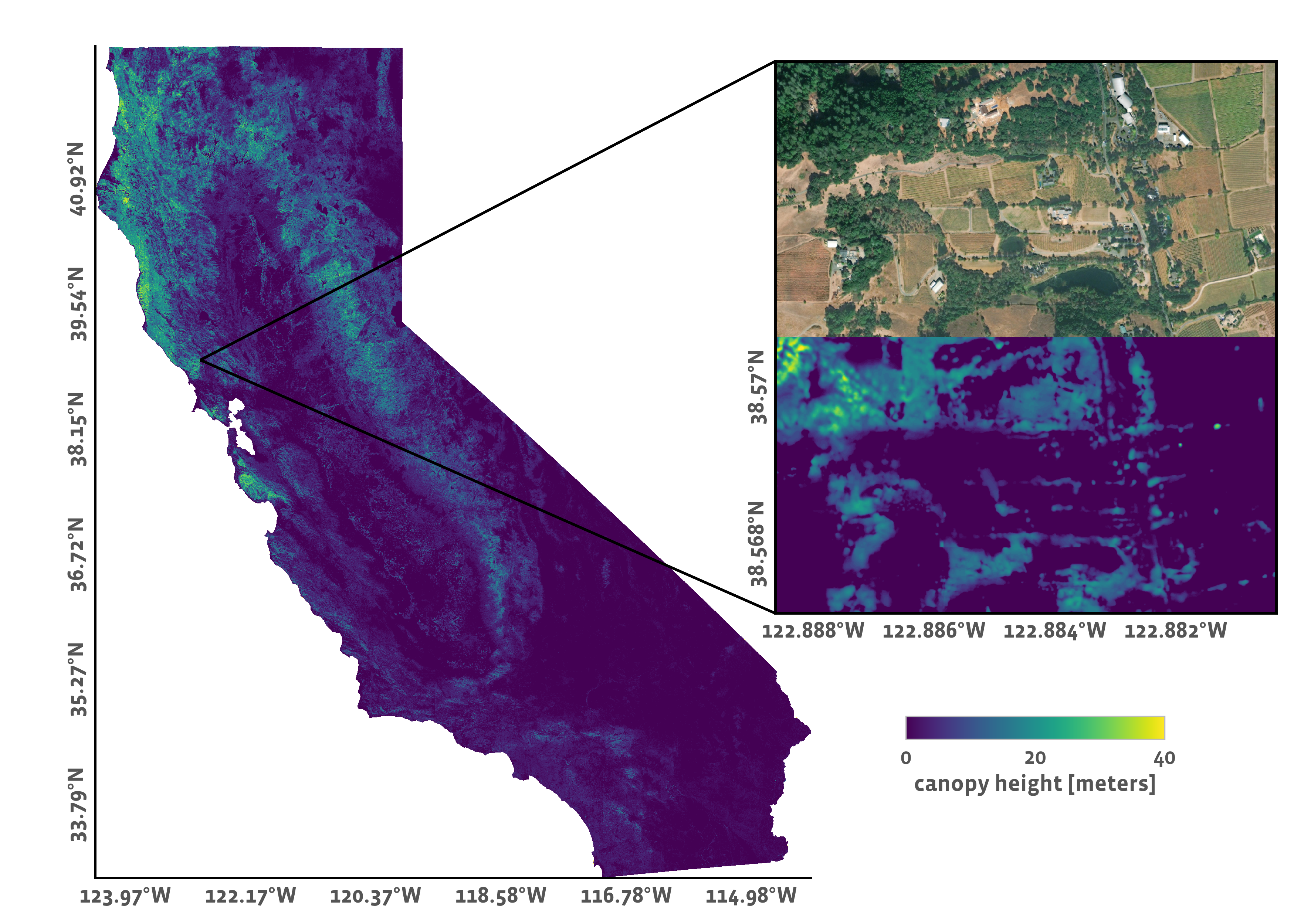}
  \caption{Canopy Height Map (CHM) for the state of California, inset showing zoomed in region with input RGB imagery.}
  \label{fig:CAoverview}
\end{figure}

\begin{figure}[ht]
  \centering 
  \includegraphics[width=1.0\textwidth]{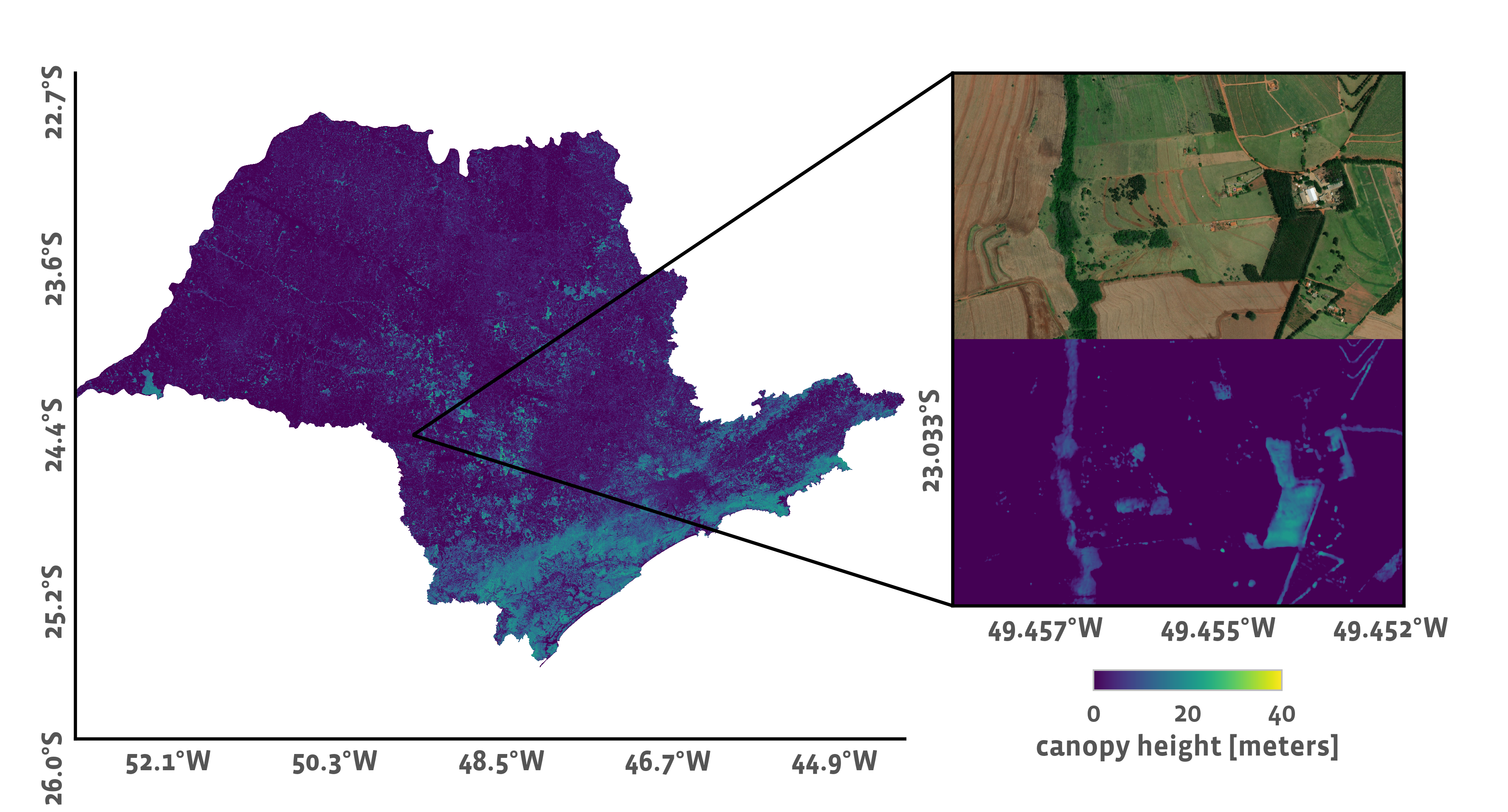}
  \caption{Canopy Height Map (CHM) for the state of S\~{a}o Paulo, inset showing zoomed in region with input RGB imagery.}
  \label{fig:SPoverview} 
\end{figure}

\begin{figure}[h]
  \centering 
  \includegraphics[width=1.0\textwidth]{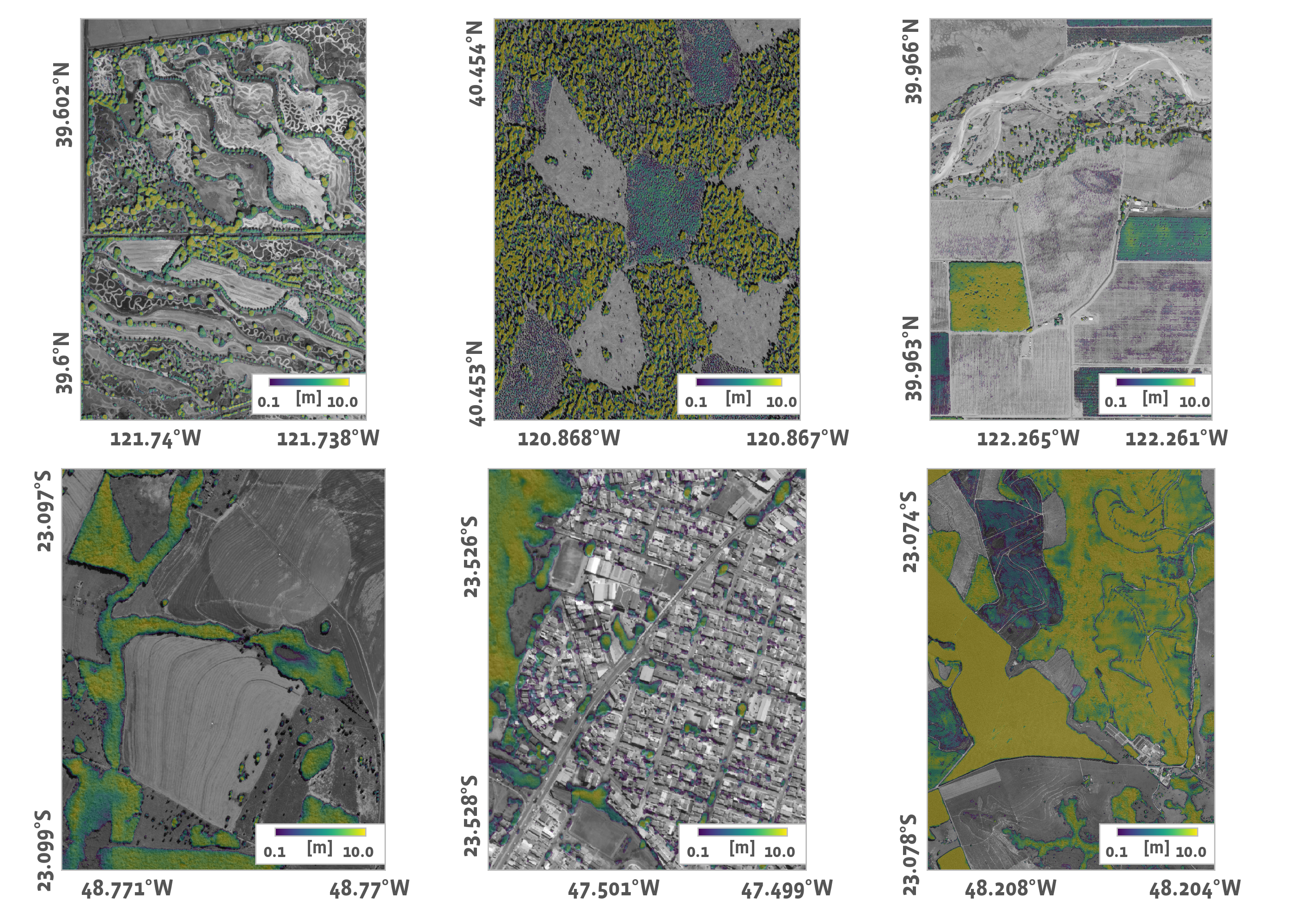}\\
  \caption{Selected sample regions from the canopy height predictions (log scale), overlayed on the input Maxar imagery (RGB). Canopy height prediction below 0.1 meter is transparent.  The top row corresponds to regions in California and the bottom row, S\~{a}o Paulo.}
  \label{fig:chmsamples}
\end{figure}

\begin{figure}[h]
  \centering 
  \begin{tabular}{ccc}
  \includegraphics[width=1.0\textwidth]{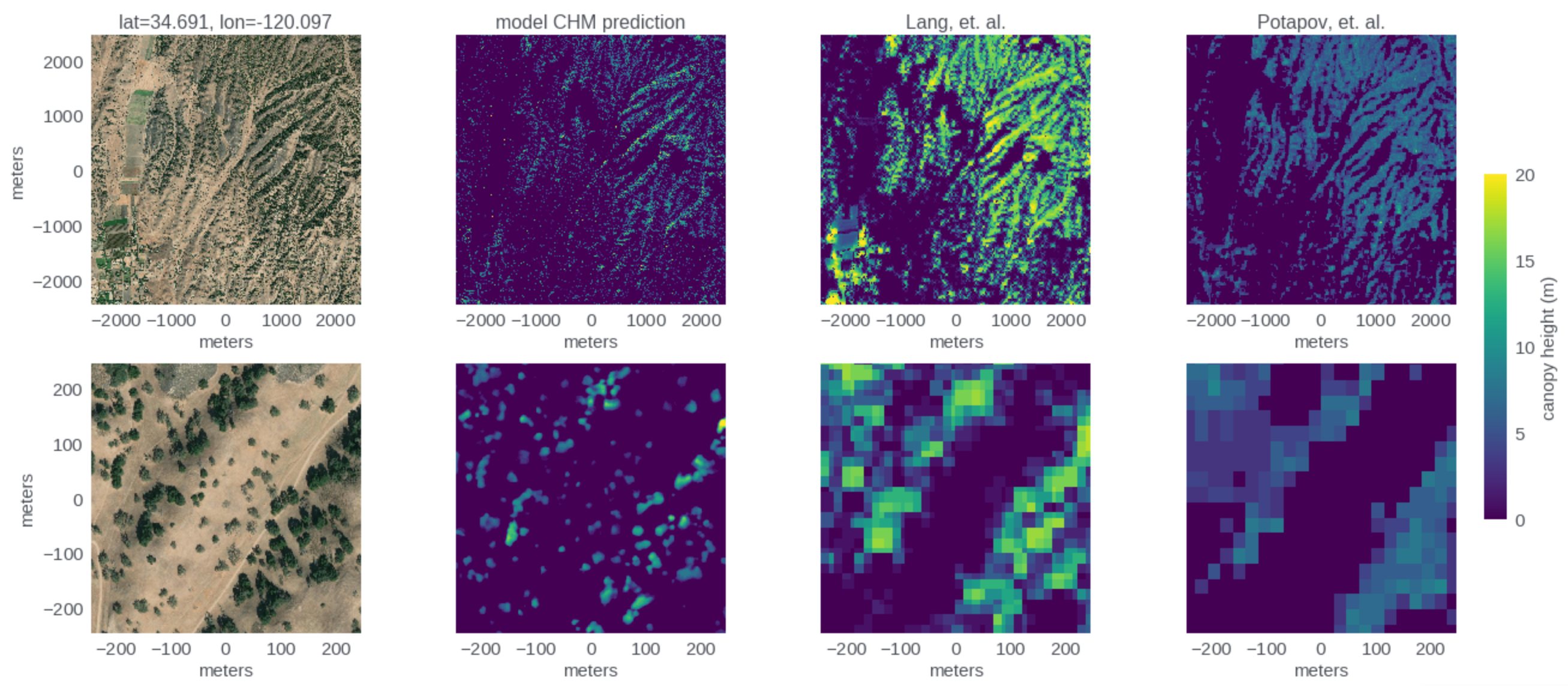}
\end{tabular}
  \caption{Comparison of our CHM (second column) with that of \cite{Lang2022High} (third column) and \cite{Potapov2021Mapping} (fourth column).}
  \label{fig:compLang}
\end{figure}

We generated CHMs for the State of California, USA (Figure \ref{fig:CAoverview}) and S\~{a}o Paulo, Brazil (Figure \ref{fig:SPoverview}) by running inference on 0.59 m GSD Maxar images with the SSL + GEDI ViT huge model trained with 1 m aerial lidar data. In California, 39 percent of the area used Maxar imagery observed in 2020, and 90 percent within the years spanning 2018 to 2020. In S\~{a}o Paulo, 63 percent of the area was observed in 2019, and 94 percent within the years spanning 2017-2019. Small regions of the canopy height predictions are visualized in Figure \ref{fig:chmsamples}. We compare our maps to the previously available highest resolution, global canopy height maps of \cite{Lang2022High} and \cite{Potapov2021Mapping} in Figure \ref{fig:compLang}. We have added the full resolution dataset to AWS Opendata programs, in the form of cloud optimized geotiffs (COGS) with associated cutlines and image acquisition dates\footnote{\url{https://registry.opendata.aws/dataforgood-fb-forests/}}. Additionally, these datasets are visible on a Google Earth Engine public url\footnote{\url{https://wri-datalab.earthengine.app/view/submeter-canopyheight}}.

\subsection{Canopy height metrics}
\label{sec:chm_metrics}
We compared the predicted canopy height maps with aerial lidar data in terms of mean absolute error (MAE), Root Mean Squared Error (RMSE), and $R^2$-block ($R^2$). The $R^2$-block score is the coefficient of determination, which we computed on cropped images with a resolution of $50 \times 50$ pixels ($\sim30\times30$ meters). We have chosen the exact size of these blocks somewhat arbitrarily, but were motivated to compute on a scale of 10s of meters due to: a.) georegistration errors in both the Maxar imagery and ALS data, b.) projection differences between the two datasets, with the ALS data being orthorectified and the Maxar imagery have off nadir view angles of up to 30 degrees. 
As such, the $R^2$-block score better reflects the local accuracy of CHMs and provides a more direct performance comparison to lower resolution models. 
However, averages across blocks of this resolution do not provide a good indicator of the edge accuracy of the produced maps, which can be a desirable property for downstream tasks such as segmentation. We separately report the Edge Error (EE) metric we developed to measure the sharpness of the maps, described in  \ref{appendix:edgeacuracy}. Finally, to estimate the bias of different models, we report the Mean Error (ME). We provide formulas for the above mentioned metrics in \ref{appendix:metrics}.

\subsubsection{Canopy height metrics for ALS models}
\label{sec:ablation}
We present in Table~\ref{tab:SSL} an ablation study of different pre-training data, model size and output on the Neon and S\~{a}o Paulo test sets. From this ablation study, we selected the SSL model trained on 18 million images utilizing the classification output, which achieved the highest canopy height accuracy metrics. We also trained a huge model instead of a large one, that significantly reduced the bias of the predictions on the S\~{a}o Paulo dataset. We refer to this model as the SSL model throughout the paper. Table~\ref{tab:SSL} suggests that pre-training on satellite images gives better results compared to pre-training on ImageNet. Compared to the ViTs that are pre-trained on ImageNet, including the SWAG approach, the ResUNet remains the strongest baseline. The SSL model clearly outperforms the ResUNet on Neon, reducing the MAE from 3.1 to 2.6 meters, is also improving results on CA Brande, and leads to similar results on S\~{a}o Paulo, with a slightly worse $R^2$ but a much lower ME.
We also experimented with different loss functions, and a smaller dataset for self-supervised pre-training. We found that was training on more data was leading to much better results in S\~{a}o Paulo. Comparing L1, L2 and Sigloss, we found that Sigloss and L2 were leading to the best results. Additional discussion of these trials can be found in \ref{appendix:loss}.

\begin{table}[ht]
\centering
\resizebox{\columnwidth}{!}{
\begin{tabular}{cccccccccccc}
\toprule
 & & pre-training & \multicolumn{3}{c}{NEON test set} & \multicolumn{3}{c}{S\~{a}o Paulo} & \multicolumn{3}{c}{CA Brande}\\ 
{} & model size &  dataset  & MAE & $R^2$-block & ME & MAE & $R^2$-block & ME& MAE & $R^2$-block & ME  \\
\midrule
ResUNet & RN18 &IN1k & 3.1 & 0.63 & {\bf 0.0}  & 5.2 & 0.42 & -2.2 & 0.6 & 0.74 & {\bf -0.1}\\
SWAG C& ViT L & IG & 3.0  & 0.63 & -1.6 & 5.8 & 0.16 & -4.3 & 0.7 & 0.56& -0.6\\
DINOv2 R& ViT L & IN1k & 3.4  & 0.52 & -1.4 & 6.8 & -0.20 & -5.2 & 0.6& 0.67& -0.4\\ 
DINOv2 R& ViT H & IN22k+ & 3.0 & 0.62 & -1.4 & 5.7 & 0.27 & -2.9 & 0.6& 0.62 & -0.4\\ 
DINOv2 R& ViT L & Sat 3.5M & 2.8 & 0.67 & -1.2 & 6.0 & 0.14  & -4.2 & 0.6& 0.70& -0.5\\ 
DINOv2 R& ViT L & Sat 18M & 2.9 & 0.66  & -1.4 & {\bf 4.9} & {\bf 0.46} & -1.9  & 0.6 & 0.68& -0.5\\
DINOv2 C& ViT L & Sat 18M & 2.7 & {\bf 0.70} & -0.9 & 5.0 & {\bf 0.46} & -2.1 & 0.6 & 0.80 & -0.3\\
DINOv2 C& ViT H & Sat 18M & {\bf 2.6} & {\bf 0.70} & -0.9 &  5.2 & 0.39 & {\bf 0.4} & 0.6 & {\bf 0.81} & {\bf -0.1}\\
\bottomrule
\end{tabular}}
\caption{Comparison of results with SSL pre-training on different datasets and with other supervised strategies (ResUNet, SWAG). IN: ImageNet. Sat: dataset described in Section~\ref{sec:ssl_data}. IG: Instagram dataset. R: DPT decoder with a regression (scalar) output. C: DPT decoder with a classification (256 bins) output. ViT L: large, H: huge. Note that the results are non GEDI corrected in this table, and all models were trained with a Sigloss. We later denote the model in the last line as the ``SSL" model.}
\label{tab:SSL}  
\end{table} %

\subsubsection{Canopy height metrics for ALS + GEDI models}

Table~\ref{tab:chm_metrics} presents a quantitative validation of the best performing models, namely the ResUNet and the self-supervised model (SSL), combined with the GEDI correction step (ResUNet+GEDI, SSL+GEDI). We note the improved performance of the SSL model compared to the ResUNet in the NEON test and CA-Brande datasets. 

Although the SSL model performed the best across the datasets in the USA (NEON test and CA-Brande), it performed worse than the ResUNet and ResUNet + GEDI for S\~{a}o Paulo, possibly due to the large domain shift in ecosystems. In the case of S\~{a}o Paulo, we found that the inclusion of GEDI (``SSL+GEDI") produced the best results, possibly indicating better generalization by including the globally trained GEDI model, which also includes additional metadata such as geographic position (Figure \ref{fig:train}).

\begin{table}[ht]
\centering
\resizebox{\columnwidth}{!}{
\begin{tabular}{rcccccccccccc} 
\toprule
&  \multicolumn{3}{c}{NEON test} & \multicolumn{3}{c}{CA-Brande} & \multicolumn{3}{c}{S\~{a}o Paulo} &\multicolumn{3}{c}{Average}\\
& MAE & RMSE & $R^2$ & MAE & RMSE & $R^2$ & MAE & RMSE & $R^2$  & MAE & RMSE & $R^2$\\
\midrule
ResUNet & 3.1   & 4.9 &   0.63	&	0.6&	1.6 &  0.75	&	5.2	& 7.4 &{\bf 0.42}	& 3.0 & 4.6 &  0.60\\
ResUNet + GEDI & 3.0    &  4.8 &   0.64	&	0.6& 1.6 &	  0.74	&	5.4	& 7.7 &  0.35	& 3.0 & 4.7 &  0.58\\
SSL & \bf{2.6} & {\bf 4.4} & \bf{0.70}	&	0.6 & {\bf 1.4} &	\bf{0.82} &	5.2	& 7.5 & 0.39	&	{\bf 2.8} & 4.5 &	\bf{0.64}\\ 
SSL + GEDI & 2.7 & 4.5 & 0.69 & 0.6& 1.5 & 0.80 & \bf{5.1} & {\bf 7.3} & 0.41 &\bf {2.8}	& {\bf 4.4} & 0.63\\
\bottomrule
\end{tabular}}\\
\caption{Canopy Height Metrics to assess the gedi correction step. $R^2$ corresponds to $\sim30\times30$ meter block $R^2$. ``Average" is the unweighted average across datasets.} 
\label{tab:chm_metrics}
\end{table}

\begin{figure}[htb]
  \centering 
  \includegraphics[width=1.0\textwidth]{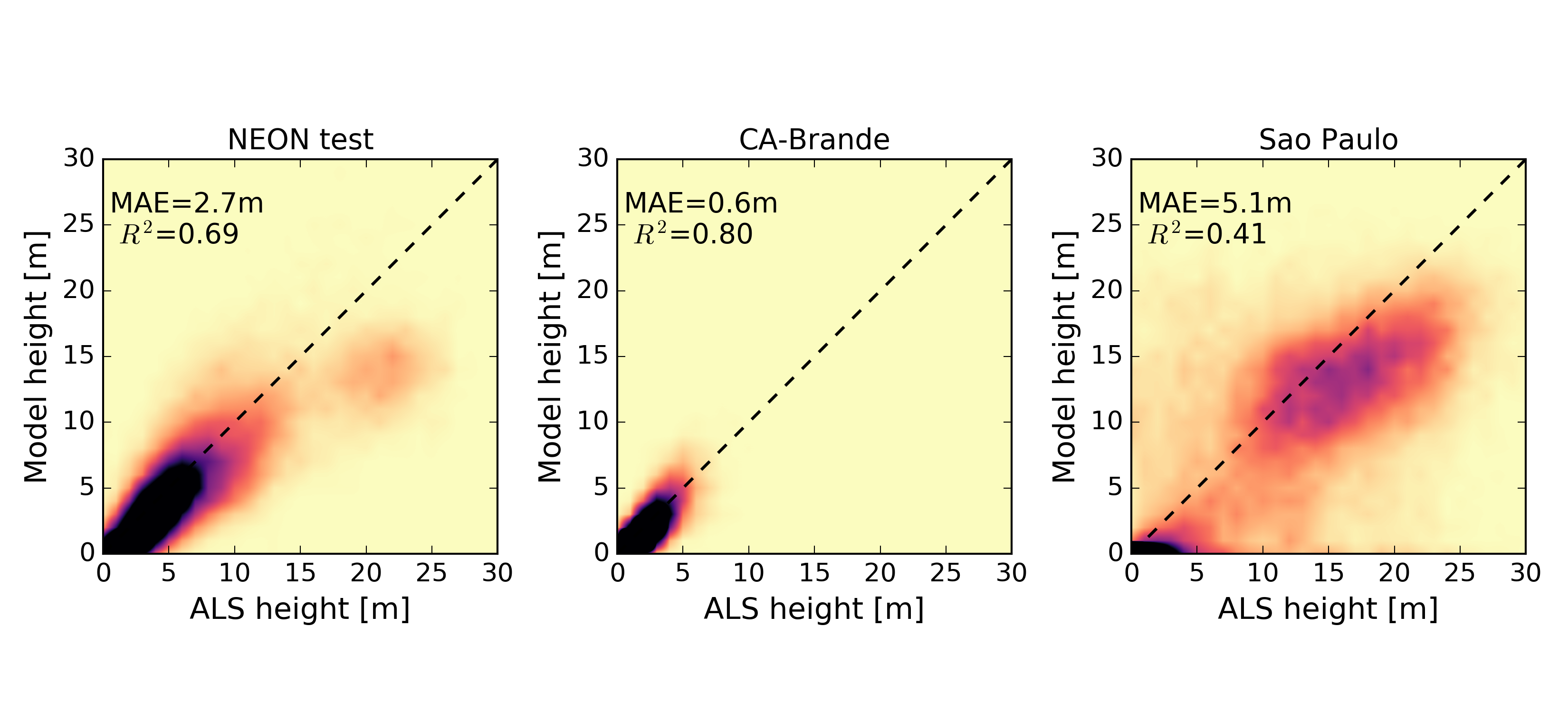}
  \caption{Block ($\sim30m\times30m$) aggregated SSL + GEDI model predictions compared to ALS ground truth measurements for different set-aside validation datasets. Colormap density is normalized to the 99.6th percentile of the heatmaps.}
  \label{fig:correlation}
\end{figure}

Figure~\ref{fig:correlation} shows 2D-histograms of the SSL+GEDI model predictions vs the set-aside validation ALS-derived canopy height averaged over $\sim30$m blocks and the corresponding pixel MAE and block-$R^2$ scores.

\subsubsection{Quantitative comparison of CHM model with GEDI RH95 data}
\label{sec:gedi_comp}

\begin{figure}[h]
  \centering
    \begin{subfigure}[b]{1.0\textwidth}
    \includegraphics[width=1\linewidth]{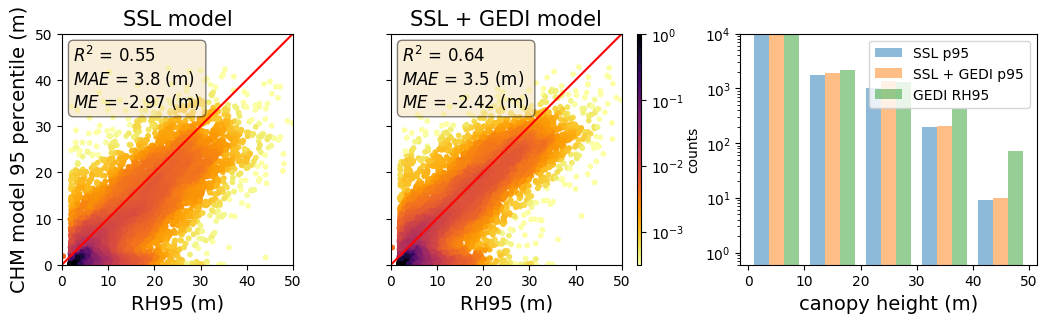} 
       \caption{MAE: mean absolute error. ME: mean error. $R^2$: Coefficient of determination.}
       \label{fig:CHM p95 vs. GEDI rh95 correlation and histogram}
    \end{subfigure}
    \begin{subfigure}[b]{1.0\textwidth}
    \begin{tabular}{cc}
       \includegraphics[width=0.75\linewidth]{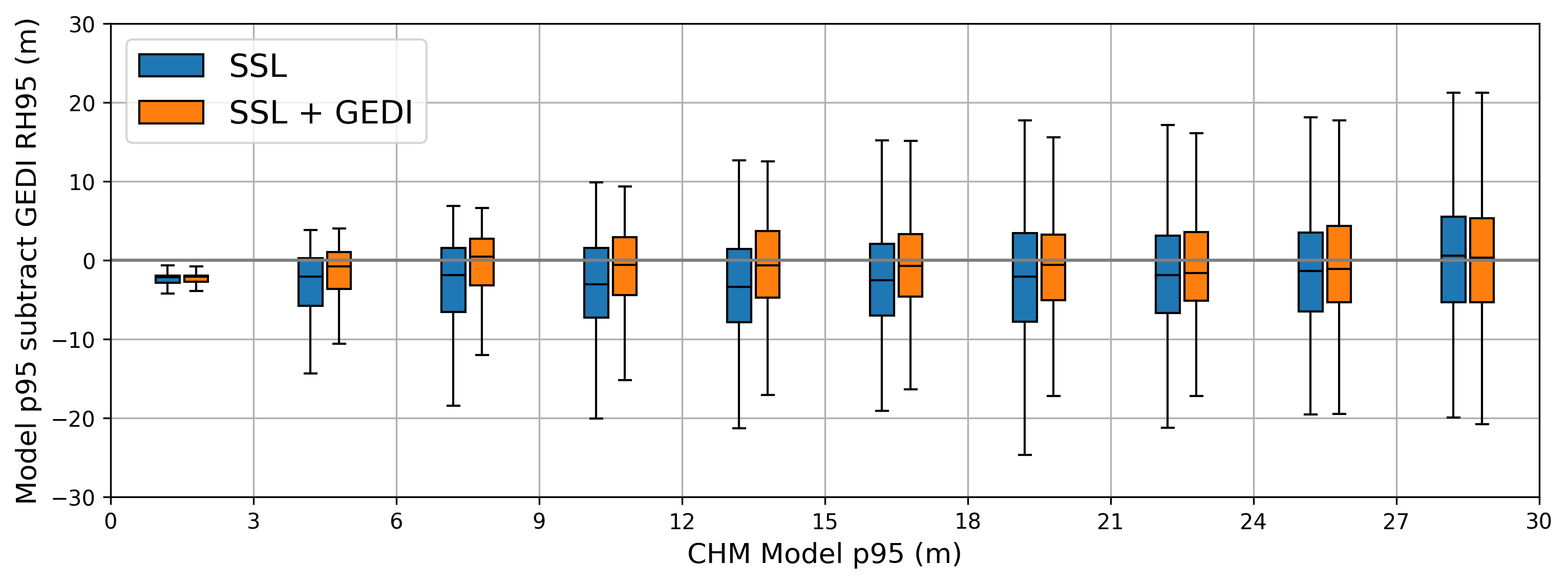}& 
         \includegraphics[width=0.25\linewidth]{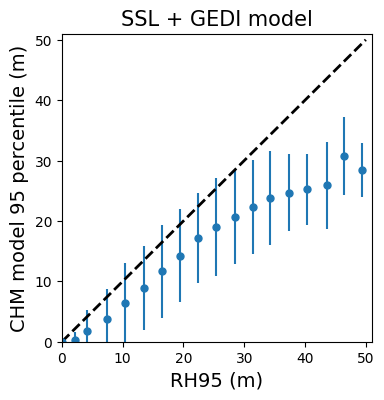}\\ 
         \end{tabular}
       \caption{}
       \label{fig:CHM p95 vs. GEDI rh95 residuals}
    \end{subfigure}
  \caption{Global model evaluation on held-out GEDI data. (a) p95 of block ($76m \times 76m$) model CHM predictions compared to the measured GEDI RH95 metrics. (b) Left: Difference between the p95 of block model CHM predictions and the measured GEDI RH95 metrics w.r.t model CHM predictions. Negative values indicate that the model estimates are lower than the GEDI RH95 values. Residuals in function of RH95 appear in \ref{app:gedi_plot}. Right: CHM p95 in function of RH95.}
\end{figure}

The ALS set-aside validation datasets used in the previous section are limited in both total area and geographic coverage. In this section, we leverage the global coverage of the GEDI dataset to validate our CHM models. As described in \ref{sec: connection between ALS P95 and GEDI RH95}, CHM maps can be connected to GEDI RH95 metrics by taking the 95th percentile. In this analysis, we draw \num{2e4} GEDI samples globally in the set-aside validation split the same way as in training the GEDI model, i.e., weighted proportional to the square root of the inverse sample size of its RH95 bin. Similarily to \cite{Potapov2021Mapping}, we removed  GEDI observations corresponding to $<$ 0.5 normalized difference vegetation index (NDVI) that had no tree cover in the 2010 data of \cite{doi:10.1126/science.1244693}, corresponding to 337 samples out of 20000. In Fig.~\ref{fig:CHM p95 vs. GEDI rh95 correlation and histogram}, we show the scatter plot and histogram of the $128\times128$ pixels ($76m \times 76m$) block $95^{\mathrm{th}}$ percentile vs. the measured GEDI RH95. In Fig.~\ref{fig:CHM p95 vs. GEDI rh95 residuals}, we analyze the difference of the CHM p95 and the GEDI RH95 with respect to the referenced GEDI RH95 heights. 

We found that the p95 of the CHM model had a small negative bias against the GEDI RH95 values and adding the GEDI correction to the CHM model significantly reduces the bias. There is a slight positive bias in the GEDI RH95 data due to the terrain slope \citep{Lang2022High}.

We used terrain slope \citep{terraintiles} as one of the input metadata when training the GEDI model (see Section~\ref{sec:gedimodel}), while setting the terrain slope to zero during inference. With this setup, we were able to calibrate out the positive bias caused by terrain slope in our GEDI model.

To assess the importance of the GEDI calibration model for geographic generalization, and the generalizability of the different baseline models, we computed R$^2$ on globally distributed GEDI test data (Table \ref{tab:gedi_geography}).

\begin{table}[htb]
\begin{center}
\begin{tabular}{lllllll}
\toprule
Subregion & \multicolumn{2}{c}{SSL}& \multicolumn{2}{c}{ResUNet}& \multicolumn{2}{c}{SWAG}\\ 
GEDI & + & - & + & - & + & - \\
\midrule
Central Asia             & 0.22 & 0.19 & \textbf{0.25} & 0.23 & 0.23 & 0.17 \\
Eastern Asia              & \textbf{0.50} & 0.44 & 0.47 & 0.42 & 0.43 & 0.38 \\
Eastern Europe              & \textbf{0.70} & 0.66 & 0.67 & 0.61 & 0.67 & 0.63 \\
Latin America + Caribbean & \textbf{0.68} & 0.64 & 0.65 & 0.56 & 0.64 & 0.56 \\
Melanesia                   & \textbf{0.52} & 0.48 & 0.51 & 0.41 & 0.44 & 0.45 \\
Northern Africa             & \textbf{0.12} & 0.11 & 0.10 & 0.06 & 0.06 & 0.05 \\
Northern America            & \textbf{0.73} & 0.69 & 0.70 & 0.65 & 0.69 & 0.64 \\
Northern Europe             & \textbf{0.54} & 0.46 & 0.41 & 0.30 & 0.33 & 0.33 \\
Oceana             & \textbf{0.68} & 0.63 & 0.66 & 0.58 & 0.61 & 0.54 \\
South East Asia            & \textbf{0.46} & 0.36 & 0.45 & 0.34 & 0.44 & 0.32 \\
Southern Asia               &\textbf{0.52} & 0.49 & \textbf{0.52} & 0.48 & 0.47 & 0.42 \\
Southern Europe             & 0.46 & \textbf{0.47} & 0.42 & 0.37 & 0.46 & 0.40 \\
Sub-Saharan Africa          &\textbf{0.68} & 0.66 & 0.58 & 0.50 & 0.64 & 0.59 \\
Western Asia                & \textbf{0.53} & 0.49& \textbf{0.53} & 0.47 & 0.47 & 0.42 \\
Western Europe              & \textbf{0.64} & 0.59 & \textbf{0.64} & 0.55 & 0.58 & 0.50 \\
\midrule
Overall              & \textbf{0.61} & 0.52 & 0.59 & 0.44 & 0.54 & 0.37 \\
\bottomrule
\end{tabular}\\
\end{center}
\caption{$R^2$ between predicted CHM p95 and GEDI RH95 by geographic subregion for 20,000 test GEDI observations for models with and without the GEDI calibration model.}
\label{tab:gedi_geography}
\end{table}

 We found that the SSL + GEDI model had the highest agreement with GEDI RH95 data in 13 of 15 geographic regions. In 42 out of 45 combinations of subregions and models, including the GEDI correction model increased R$^2$. 

\subsubsection{Correlation with field data}
\label{sec:heightcorr}

\begin{figure}[htb]
  \centering 
  \includegraphics[width=0.5\textwidth]{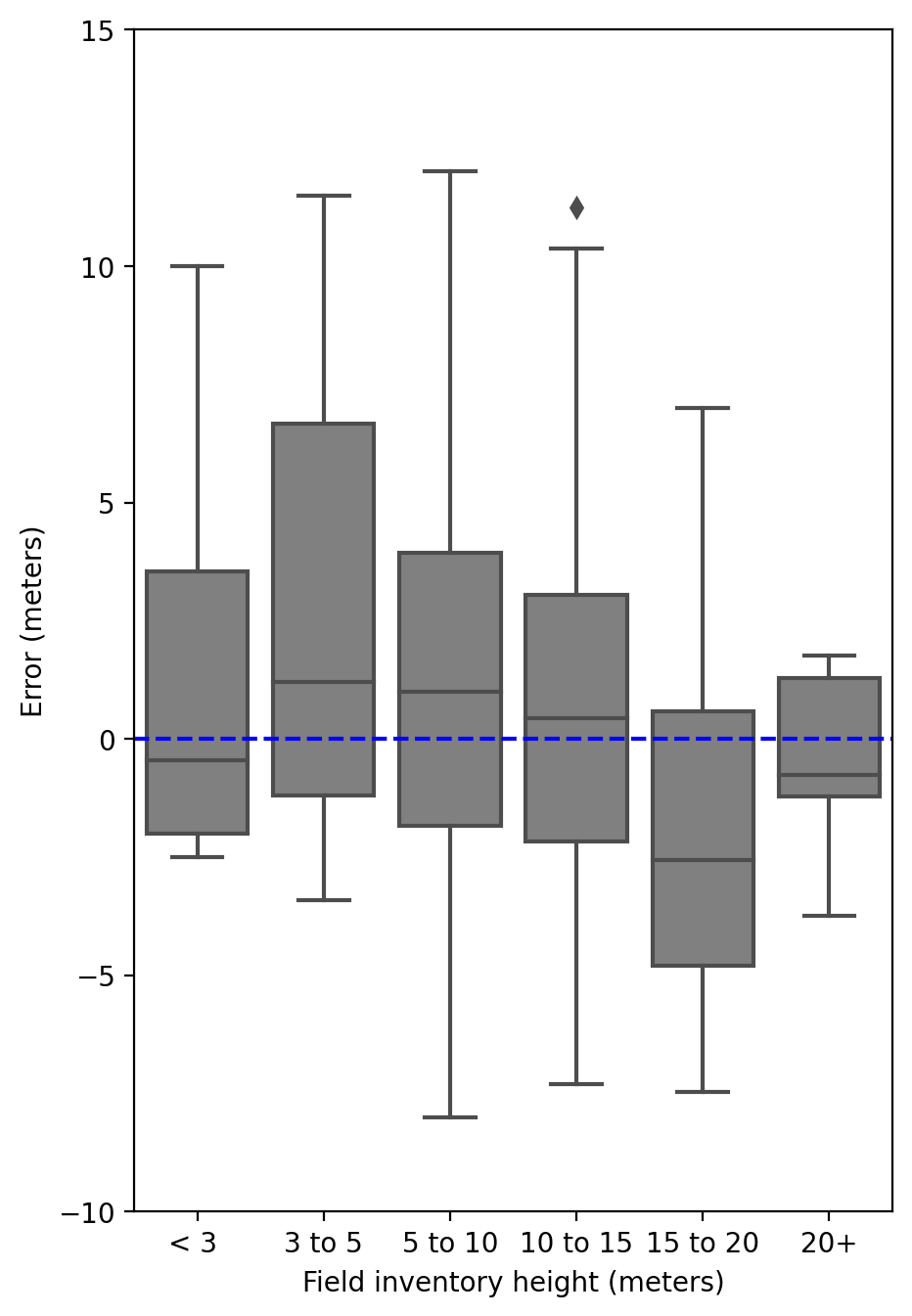}
  \caption{CHM error compared to reference tree height as indicated in the Brazilian National Forest Inventory for Espirito Santo.}
  \label{fig:NFI}
\end{figure}

To measure the agreement between our computed CHMs and field-collected tree height data, we utilize the Brazilian National Forest Inventory (NFI) data, which consists of systematic field plot inventories of tree count and height \citep{brazilnfi}. Because the NFI data for S\~{a}o Paulo was not yet available, we additionally generate a CHM of the nearby Espirito Santo state and evaluate its agreement with the NFI data for Espirito Santo. The NFI data analyzed encompassed 1,450 10$\times$10 m subplots within 87 plots positioned within a 20$\times$20 km grid in Espirito Santo. The field data was collected primarily in November and December 2014, and includes the height of each tree within each subplot having a diameter at breast height (DBH) of at least 10 cm. Of the 1,450 initial plots considered, we removed 291 that had tree cover loss since 2014 in the dataset of \cite{Hansen_2016}. Figure \ref{fig:NFI} visualizes box plots of the 95th percentile CHM by reference NFI height bins. The overall ME was 0.72 m while the RMSE was 4.25 m, with a slight positive bias for trees $\le$15 m (ME = 1.10 m, RMSE = 4.28 m), and negative bias for trees $>$15 m (ME = -1.00 m, RMSE = 3.79 m).

\subsection{Segmentation metrics}
\label{sec:seg_metrics}
In addition to the canopy height metrics discussed in Section \ref{sec:chm_metrics}, we compute a number of metrics that reflect the ability of the model to correctly assign individual pixels as trees. CHMs were converted into binary masks by thresholding height values of at least five meters (5m) as tree canopy extent.  Table \ref{tab:seg_metrics} shows the pixel user's and producer's accuracy values (also know as precision and recall, respectively) for pixels labeled as trees. Table \ref{tab:seg_metrics} also shows the Intersection Over Union (IOU) for the binary masks, which was calculated as the average of IOU for pixels labeled as tree and the IOU for pixels labeled as ground. 

Additionally, we introduce an Edge Error (EE) metric that computes the ratio of the sum of the absolute difference between edges from predicted and ground truth CHM, normalized by the sum of detected edges in both maps. Scores range between 0 and 1, where lower scores indicate improved accuracy along patch edges. In Table~\ref{tab:seg_metrics}, the edge error is computed over all set-aside validation datasets. We detail the formula with a figure illustrating the behavior of this metric in \ref{appendix:edgeacuracy}.

\begin{table}[ht]
\centering
\resizebox{\columnwidth}{!}{
\begin{tabular}{rccccccccccc} 
\toprule
&  \multicolumn{2}{c}{NEON test} & \multicolumn{2}{c}{CA-Brande} & \multicolumn{2}{c}{S\~{a}o Paulo}  & \multicolumn{3}{c}{Average}\\
& U/P &  IOU & U/P & IOU & U/P & IOU & U/P & IOU & EE\\
\midrule
ResUNet & 0.74/0.75   &  0.58	&	0.72/0.64&	0.70 &	{\bf 0.91}/0.85	& {\bf 0.67}	& 0.79/0.75& 0.65 & 0.50 \\
ResUNet + GEDI & 0.77/0.68   & 0.53	&	0.73/0.52&	0.68 &	{\bf 0.91}/0.84	& 0.65	& 0.80/0.68& 0.62 & 0.52 \\
SSL & 0.81/{\bf0.76}	&	{\bf 0.65} &	0.71/{\bf 0.75}	&	{\bf 0.76} & 0.90/{\bf 0.88}	& 	\bf{0.67}  & 0.82/{\bf 0.81} & {\bf0.68} & 0.50\\ 
SSL + GEDI & {\bf 0.82}/0.71 & 0.59 &  {\bf 0.74}/0.60 &  0.74 & {\bf 0.91}/0.86 & 0.66 & {\bf 0.83}/0.76& 0.66 &\bf{0.49} \\
\bottomrule
\end{tabular}}\\
\caption{Segmentation metrics. U/P corresponds to pixel user's / producer's accuracy of the tree class. IOU to the average of tree $\&$ no tree IOU class scores. EE: Edge error. }
\label{tab:seg_metrics}
\end{table}

We observe an improvement of the SSL approach over the ResUNet baseline in terms all segmentation metrics. Both approaches leads to maps with the same level of sharpness, and the GEDI correction slightly degrades results.

\subsubsection{Tree detection metrics evaluated against human annotated validation data}
\label{sec:man_annot}

To assess the ability of the model to generalize to new geographies, we compiled human-annotated validation labels for tree detection (binary classification of tree vs no-tree) across $8,903$ Maxar thumbnail images. Human annotators were instructed to label any trees above one meter (1m) tall and with a canopy diameter of more than three meters (3m). Annotators were to include standing dead trees and tree-like shrubs, but exclude any grasslands, row crops, lawns, or otherwise vegetative ground cover whose peak height was estimated to be less than 1m from the ground surface. To create the model binary masks for the annotated dataset, we thresholded the model CHM at 1m. 

\begin{figure}[ht]
  \centering 
  \includegraphics[width=1.0\textwidth]{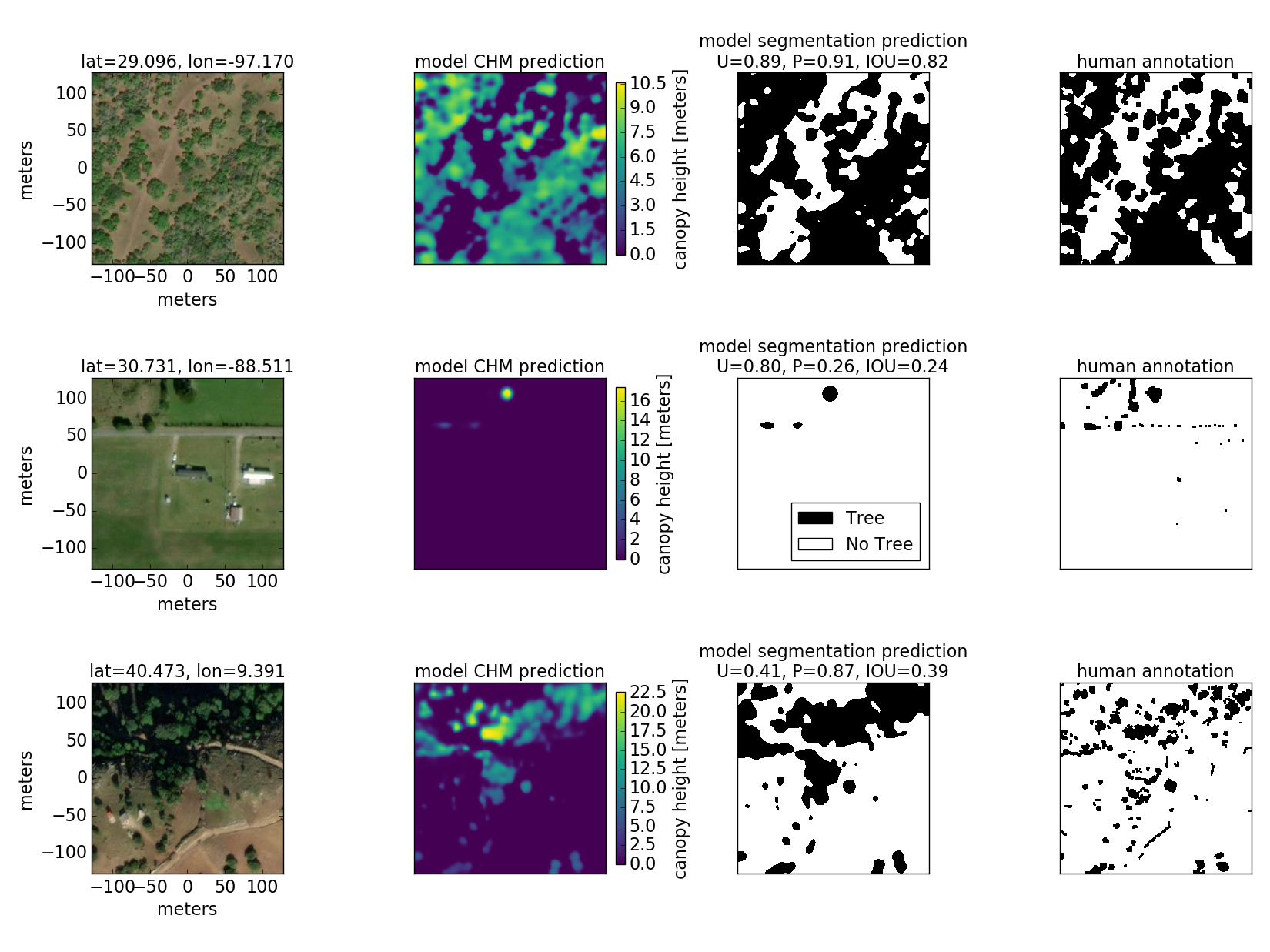}
  \caption{Tree segmentation predictions from the SSL + GEDI model vs human annotated ground truth. Binary prediction masks were created from the CHM by thresholding at 1m. U/P corresponds to pixel user's / producer's accuracy of the tree class. The IOU represents the Intersection-Over-Union score for the tree class.}
  \label{fig:tree_segmentation}
\end{figure}

The geographic locations for the images in the dataset correspond to randomly sampled GEDI measurement footprints from our global set-aside validation set where the GEDI measurement had RH95 greater than 3 meters, which we enforce to bias the dataset towards vegetated areas. The data is independent of the aerial lidar measurements used to train the model. Over the entire dataset, the user's and producer's accuracy was $0.88\pm0.006$ and $0.82\pm0.008$, while the IOU was $0.77\pm0.006$ indicating good agreement with the human annotations, cf. Table~\ref{tab:seg_metrics2}. Figure~\ref{fig:tree_segmentation} shows examples of model predictions and their corresponding annotations.

\begin{table}[ht]
\centering
\begin{tabular}{rcc} 
\toprule
& \multicolumn{2}{c}{Global, Annotated}\\
& U/P & IOU\\
\midrule
ResUNet  & 0.89/0.86	&  0.75\\
ResUNet + GEDI & {\bf0.90}/0.86	&  0.74\\
SSL & 0.83/0.87 & \bf{0.77} \\ 
SSL + GEDI &  0.82/\bf{0.88}	& \bf{0.77}\\
\bottomrule
\end{tabular}\\
\caption{Segmentation metrics on global, human annotated dataset. U/P corresponds to pixel user's / producer's accuracy. IOU to the average of tree $\&$ no tree IOU scores. Since the GEDI correction only adjusts large scale height percentiles, the ``+GEDI'' rows show only small improvement over the base ALS models.}
\label{tab:seg_metrics2}
\end{table}

\begin{figure}[htb]
  \centering 
  \includegraphics[width=0.9\textwidth]{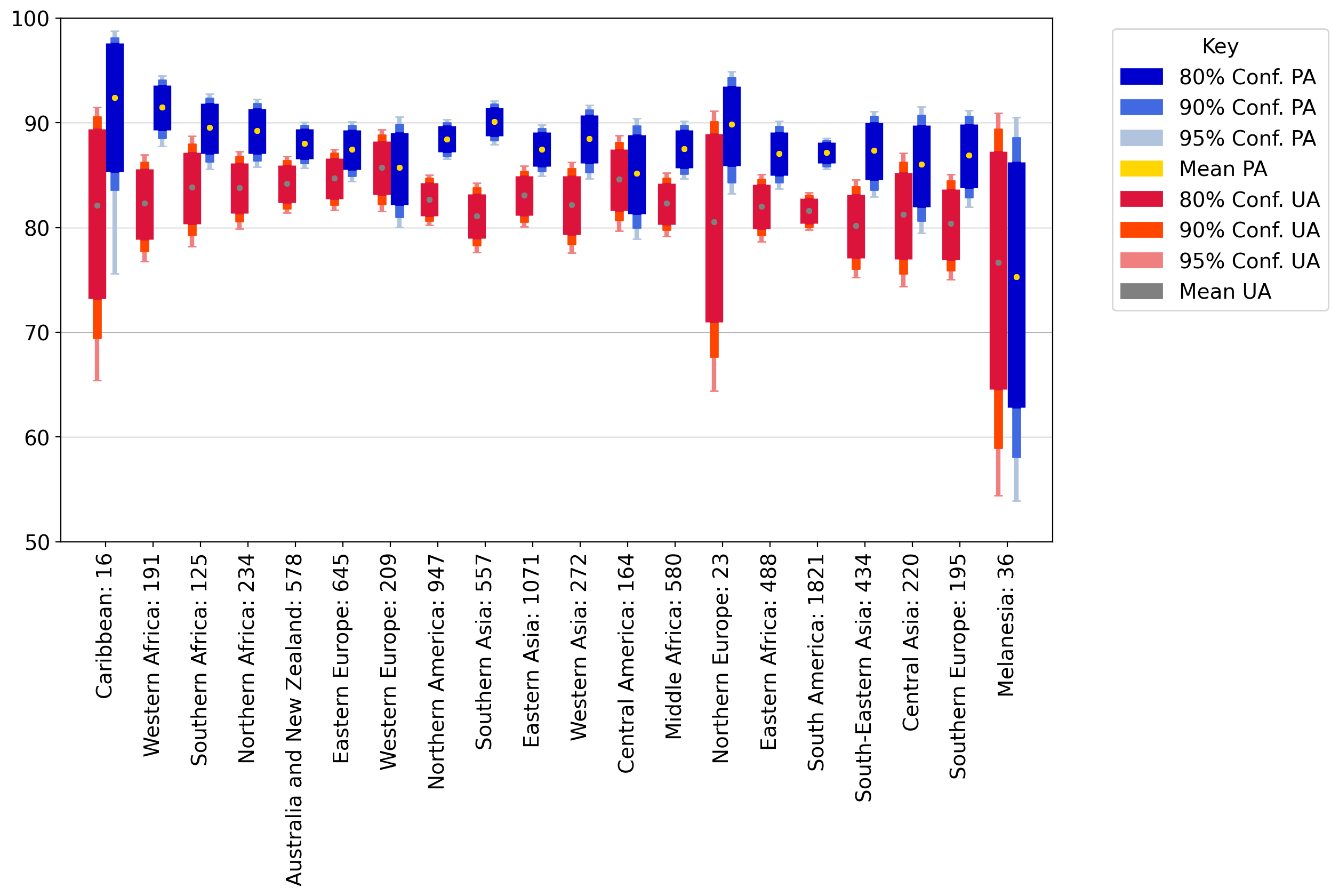}
  \caption{Pixelwise user's accuracy (UA) and producer's accuracy (PA) for 8,903 validation plots stratified by geographic sub-region. Error bars represent the 80, 90, and 95 percent confidence intervals as derived from 10,000 bootstrap iterations. Numbers in the x-axis tick labels denote sample size.}
  \label{fig:validation_subregion}
\end{figure}

We additionally calculated user's and producer's accuracy by geographic subregion according to the United Nations geoscheme. Boostrapping with 10,000 iterations was used to calculate uncertainty for tree segmentation accuracy metrics rather than the methods of \cite{Stehman_2014} because the cluster sampling approach was used to generate validation data \citep{OLOFSSON201442,mugabowindekwe2022nation,rs13132591}. This validation analysis indicated strong generalizability across different geographic regions, without significantly different accuracy metrics in geographic regions where we had training data and where we did not (Figure~\ref{fig:validation_subregion}). This suggests that the use of self supervised learning  on global images facilitated the creation of a generalizable segmentation network.  

\begin{figure}[htb]
  \centering 
  \includegraphics[width=1.0\textwidth]{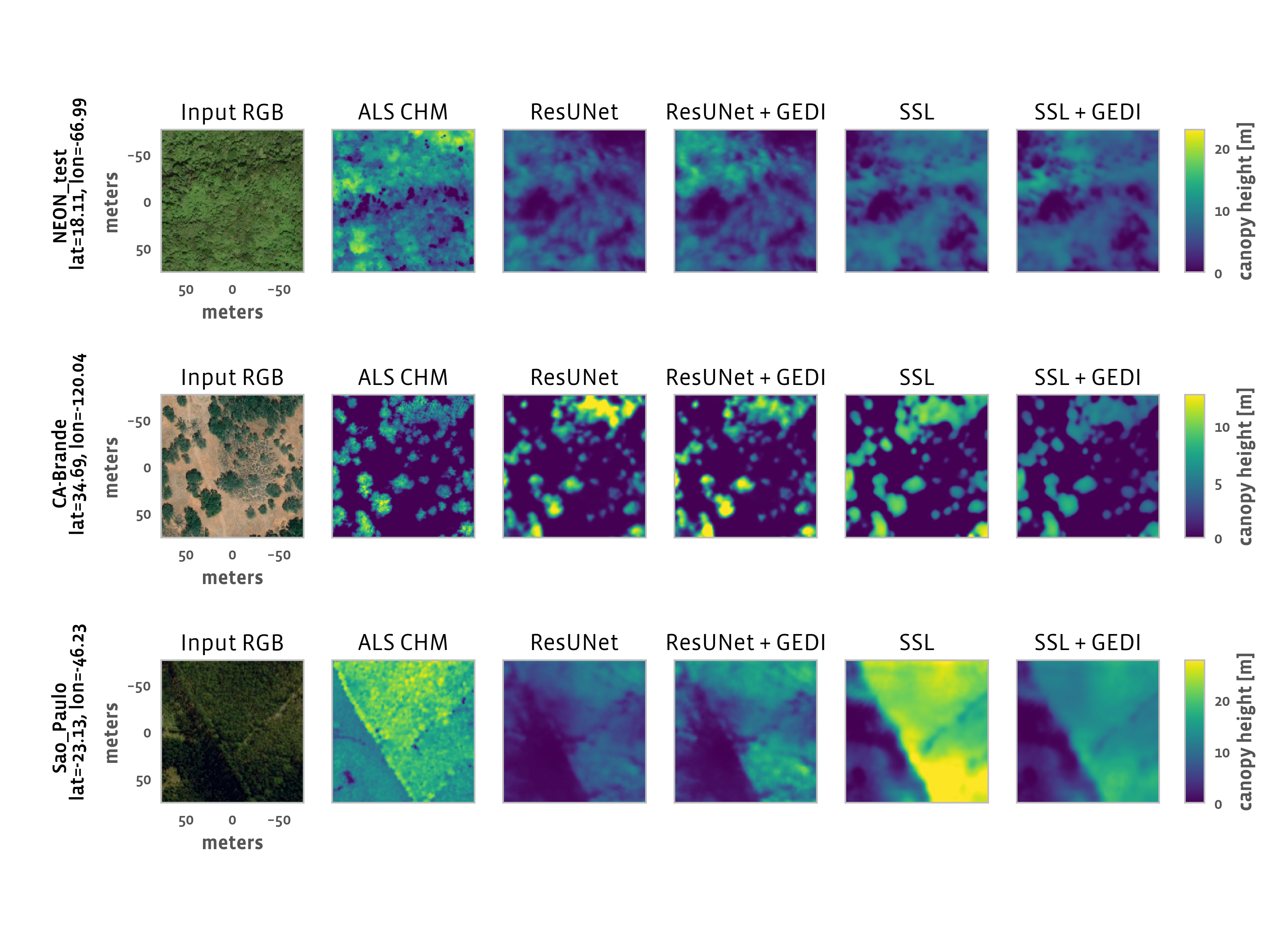}\\
  \caption{Qualitative comparison between different models for example imagery. Left: Input Maxar ``thumbnail" image, $256\times256$ pixels, in local tangent plane coordinate system. Second from left: ALS CHM image, in same projection and pixelization. Right columns: Model CHMs.}
  \label{fig:qualcomparison}
\end{figure}

 \subsection{Qualitative comparison of models}

 Although we have attempted to capture the performance of each model qualitatively with the included metrics, we note that visual inspection often leads to additional insights. Therefore, we additionally present a few examples of maps produced by our various models. Figure~\ref{fig:qualcomparison} compares the results with a ResUNet and SSL based strategies.

\subsection{Canopy height as a function of plantation age}
\label{sec:agecorr}

\begin{figure}[htb]
  \centering 
  \includegraphics[width=0.5\textwidth]{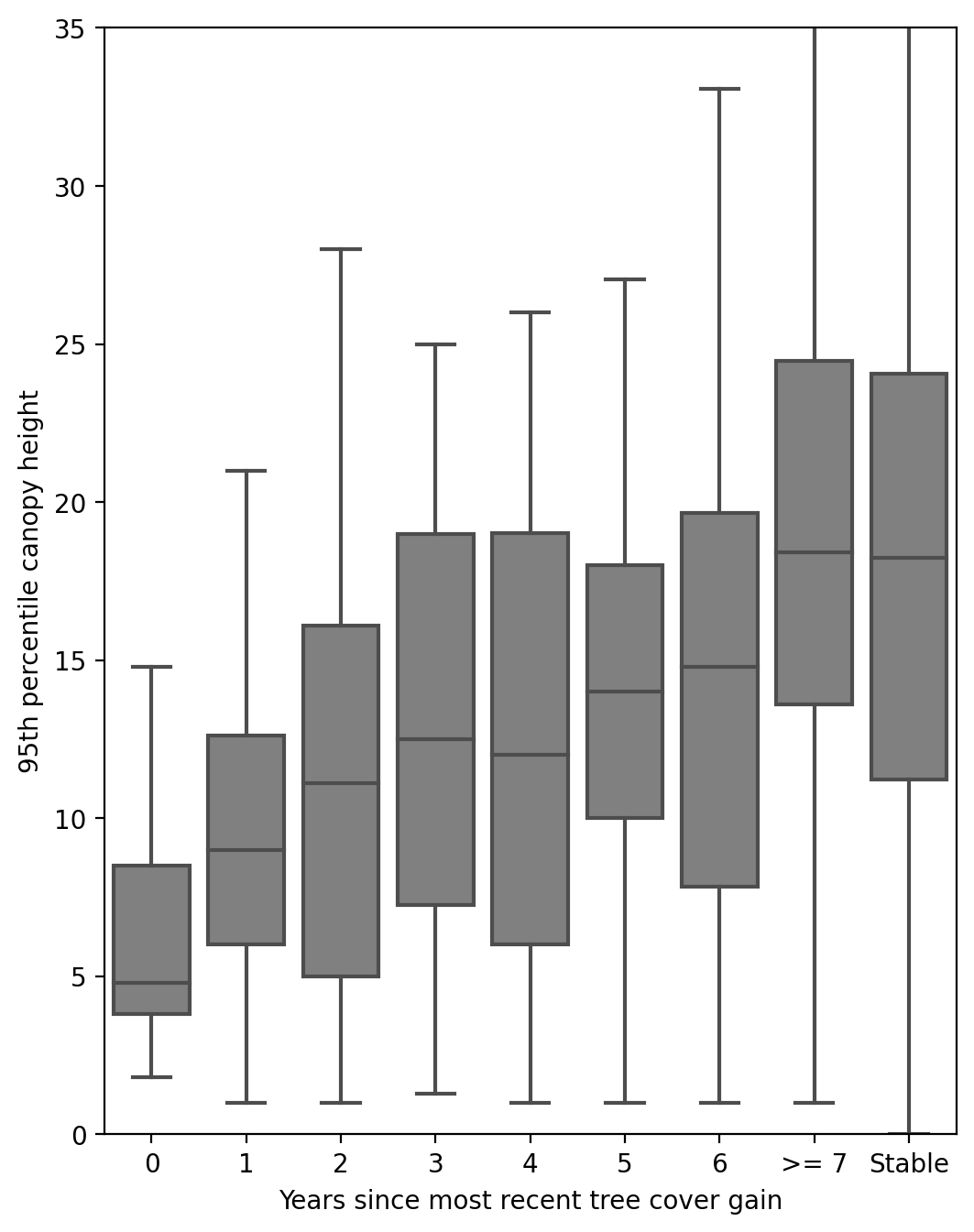}
  \caption{Canopy height estimates for areas with tree cover gain of various ages in S\~{a}o Paulo relative to the imagery year analyzed. }
  \label{fig:agecorr}
\end{figure}
Densely planted monoculture stands, such as those commonly found in the Atlantic forest, can be many hundreds of hectares large. Assessing the age-height relationship of tree stands with CHMs derived from optical imagery may be challenging due to the relative homogeneity of the canopy structures, the large area to perimeter ratio, and the lack of canopy gaps. To assess the CHMs ability to map the height of planted trees, we utilized the annual 30-meter tree cover gain and loss data from MapBiomas in S\~{a}o Paulo \citep{mapbiomas}. We calculated the number of years since the most recent tree cover gain with no subsequent loss event for each image date analyzed. Figure \ref{fig:agecorr} shows a positive relationship (R$^2$ = 0.59) between the number of years since the most recent tree cover gain, and our predicted 95th percentile CHM. For areas with gain events older than seven years, there was no significant age-height relationship, as areas with trees with gain events more than seven years prior to the analysis year had similar height distributions to areas with stable (no gain or loss since 2000) trees. For this analysis, it's important to note that the tree cover gain year identified in MapBiomas is a lagging indicator of the tree age, since tree cover gain is not immediately recognizable from Landsat imagery.

\subsection{Generalization to aerial imagery}  

\paragraph{Using a model fully trained on Satellite images}
To assess the generalization ability of our approach to other input imagery, we measure model performance using airborne imagery at inference. 
For inference, we resized the NEON aerial images to match the size of corresponding satellite image, and apply a normalization of the aerial image to match the color histogram of the satellite imagery. Details about image normalization are provided in \ref{appendix:normalization}.

\begin{figure}[htb]
  \centering 
  \includegraphics[width=1.0\textwidth]{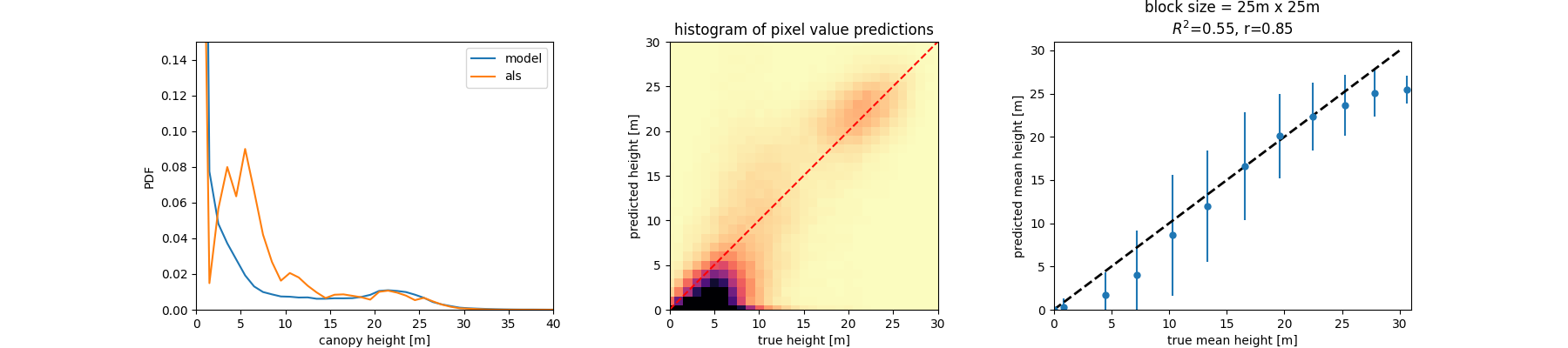}\\
\includegraphics[width=1.0\textwidth]{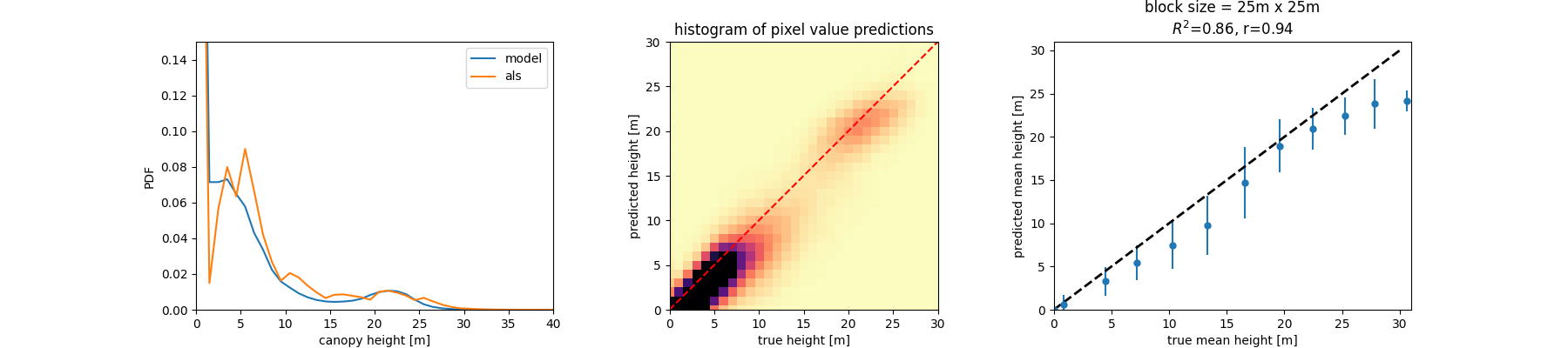}\\
  \caption{Performance of models given aerial images inputs. Top: Model fully trained on satellite images; Bottom: Performance of encoder trained on satellite images, decoder trained on aerial images.}
  \label{fig:aerial_plots}
\end{figure}

\begin{figure}[htb]
  \centering 
  \includegraphics[width=1.0\textwidth]{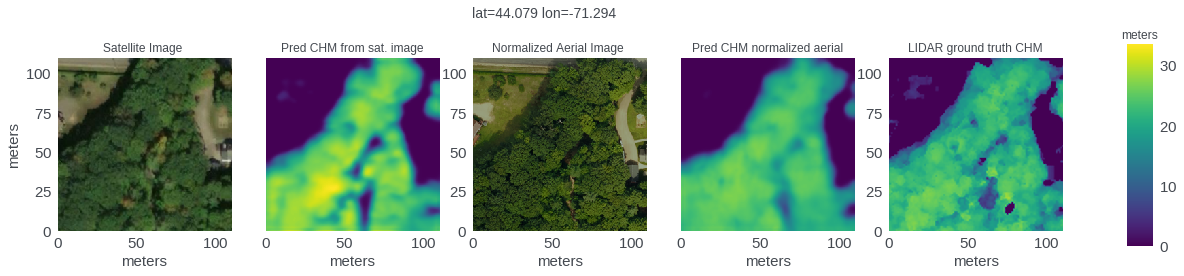}\\
  \caption{Generalisation of our SSL approach. Even if trained on Satellite images, inference on airborne images does not seem to suffer from a domain shift.}
  \label{fig:generalisation}
\end{figure}

The second line of Table \ref{tab:generalisation} shows canopy height metrics computed on predictions made from NEON input RGB imagery. The SSL model almost doubles the $R^2$ of the ResUNet baseline. Compared to the performance of the SSL model with satellite images as input as reported in Table \ref{tab:SSL}, the MAE is only slightly higher (3.0 instead of 2.7), the $R^2$ is a bit more impacted (0.55 instead of 0.70), while the bias is much higher, but evenly distributed between different height bins (Figure \ref{fig:aerial_plots}).
Figure \ref{fig:generalisation} displays a qualitative example, where we observe that despite a blurrier result, the accuracy of the model given an out-of-domain aerial image seems similar to the one obtained using in domain satellite imagery.

\begin{table}[ht]
\centering
\resizebox{\columnwidth}{!}{
\begin{tabular}{ccccccc}
\toprule
 & \multicolumn{6}{c}{Neon test - aerial} \\ 
  & Encoder training dataset & Decoder training dataset & MAE & block $R^2$ & ME & EE \\
\midrule 
ResUNet & INet & Sat. images & 3.7 & 0.34 & -2.0 & 0.77\\ 
SSL & Sat. images  & Sat. images & 3.0 &  0.55 & 1.7 & 0.71\\
SSL & Sat. images  & aerial  & {\bf 1.8} & {\bf 0.86} & {\bf -1.0} & {\bf 0.41} \\ 
\bottomrule 
\end{tabular}
}
\caption{CHM prediction accuracy on NEON test dataset using aerial input images as inputs. Trained on satellite images only, the SSL approaches demonstrates generalization abilities.}
\label{tab:generalisation}
\end{table} 

 Despite changes in color intensity, image angle, and sun angle, our approach manages to generate predictions with consistent visual quality. From an application point of view, the robustness of SSL predictions without the need to retrain models on new lidar datasets is very interesting. 

\begin{figure}[ht]
  \centering 
  \begin{tabular}{cccc}
  \includegraphics[height=0.22\textwidth]{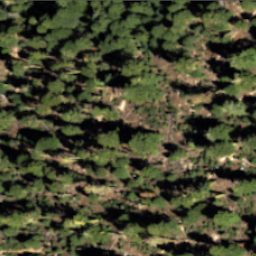}&
  \includegraphics[height=0.22\textwidth]{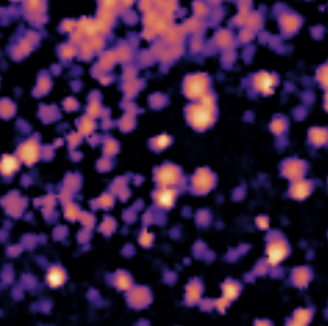}&
  \includegraphics[height=0.22\textwidth]{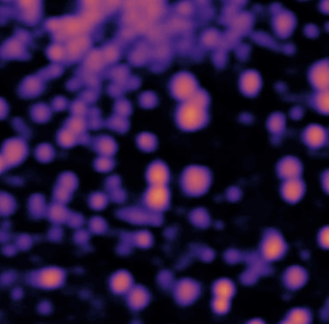}&
  \includegraphics[height=0.22\textwidth]{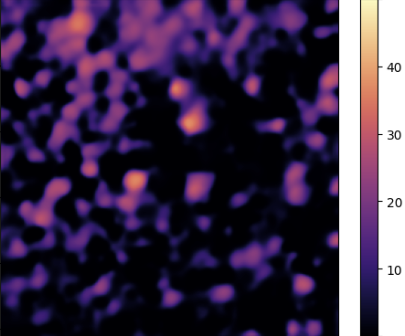}\\
  {\small RGB aerial image} & {\small Lidar Ground Truth} & {\small \cite{wagner2023submeter}} & {\small Our predicted CHM}\\
  \end{tabular}
  \caption{Comparison of our aerial model, where we trained the DPT decoder on Neon aerial RGB images, with the approach of \cite{wagner2023submeter}. Note that despite a slight change in the scale of the input image, which was zoomed to obtain a $256 \times 256$ input, and despite the fact we did not use the infra-red input, we obtain a result similar to the one of \cite{wagner2023submeter}.}
  \label{fig:comp_wagner}
\end{figure}

\paragraph{Training a new decoder on aerial images}
We compared these results to another baseline, training a decoder on top of our pretrained SSL features on Neon RGBs (last line of Table \ref{tab:generalisation}). Given a better alignment with the ground truth CHMs, and view angles close to Nadir across the Neon dataset, this aerial model performed reasonably well compared to  the recent result of  \cite{wagner2023submeter}, only using the RGB channels, as illustrated in Figure~\ref{fig:comp_wagner}.
      
\section{Discussion}

Our proposed method provides a novel approach to estimating canopy height from VHR satellite imagery. We demonstrate the effectiveness of our approach based on self-supervised learning, dense vision transformers, and introduce an approach to  rescale high-resolution canopy height maps from a model trained on Maxar and ALS data with low-resolution canopy height maps from a model trained on Maxar and GEDI data. In contrast to \cite{Lang2022High}, which downscales the 25-meter GEDI data to generate 10-meter canopy height maps by only considering Sentinel-2 pixels at the centroid of each GEDI pixel, our approach uses a GEDI-based canopy height map to rescale an ALS-based model of canopy height map. While both \cite{Lang2022High} and \cite{Potapov2021Mapping} only utilize ALS data to validate their generated maps, we directly model the relationship between Maxar imagery and ALS data, enabling the mapping of sub-GEDI scale canopy height variability, some times at a per-tree level outside of dense, closed-canopy forests.

\paragraph{Segmentation}

Previous research applying deep learning image segmentation approaches to map trees in high-resolution imagery, such as \cite{Brandt2020Sahel} and \cite{mugabowindekwe2022nation} have utilized a U-Net model \citep{ronneberger2015u} and entirely hand-labeled reference data. Focusing in Rwanda, \cite{mugabowindekwe2022nation} map carbon stock estimates for individual trees by developing empirical relationships between crown area and carbon, finding that half of the national tree carbon stock is located outside of natural forests. In comparison to these approaches, our results suggest that incorporating SSL can improve model generalizability for vegetation structure mapping, in line with various research demonstrating the effectiveness of SSL in other domains.  Additionally, our per-pixel height predictions combine the predictive quality of height for assessing biomass as demonstrated in \cite{LANG2022112760} and \cite{Potapov2021Mapping} with the predictive quality of crown area for assessing biomass as demonstrated in \cite{mugabowindekwe2022nation} and \cite{f12121652}.

\paragraph{Limitations}

The production of high-resolution canopy height maps from optical imagery has challenges and associated limitations. Foremost, the availability of recent ALS training data is limited in geographic scope. While the utilization of self-supervised learning and the GEDI-based corrective model improve generalization and reduce test error, increased geographic availability of ALS remains necessary to further validate the proposed maps in new geographies. While we were able to validate error as a function of canopy height for trees under 25 m based on field inventory data in Espirito Santo, Brazil (Figure \ref{fig:NFI}), we were unable to utilize field data to assess potential height saturation for very tall trees which may affect derived above ground carbon data. However, Figure \ref{fig:CHM p95 vs. GEDI rh95 correlation and histogram} does suggest significant saturation of predictions for GEDI RH95 observations above 30 m.

The generated maps are limited by variation in input imagery, particularly by variation in view angle, sun angle, acquisition times and dates, and optical aerosol. As shown in Figure \ref{fig:comp_wagner}, qualitative data quality improves considerably when processed on VHR aerial optical imagery, as opposed to VHR satellite optical imagery. Additionally, terrain slope appears to influence predicted height, since it affects the length of shadow an individual tree casts. At present, the ability to conduct tree height change detection assessments is limited by the need for improved input image processing to better align these differences between image pairs.

\section{Conclusion}
This study presents high-resolution canopy height maps at a jurisdictionial scale based on VHR (Maxar) optical imagery trained  on aerial lidar and calibrated with spaceborn lidar (GEDI) data using latest advances from self-supervised learning and vision transformers. We demonstrate quantitatively and qualitatively the advantages of large-scale self-supervised learning, the versatility of obtained representations allowing generalization to different geographic regions and input imagery. Compared to existing canopy height maps, the presented data better captures tree structure variability at small spatial scales. Such very high resolution maps of canopy height can improve the monitoring of forest degradation, restoration, and forest carbon dynamics. The next steps include (a) developing and validating allometrically-derived high-resolution woody carbon data and (b) testing and validating the utility of the proposed approach for the operation monitoring of tree growth. 

\section{Acknowledgments}

We would like to thank Ben Weinstein for helpful discussions regarding the NEON dataset. We thank Andi Gros and Saikat Basu for their technical advice. We would like to thank Shmulik Eisenmann, Patrick Louis, Lee Francisco, Leah Harwell, Eric Alamillo, Sylvia Lee, Patrick Nease, Alex Pompe and Shawn Mcguire for their project support. 

\appendix

\section{Data used in training / calibration / validation}
\label{appendix:neonsites}

\begin{figure}[htb]
  \centering
    \begin{subfigure}[b]{.5\textwidth}
        \centering
       \includegraphics[width=0.85\linewidth]{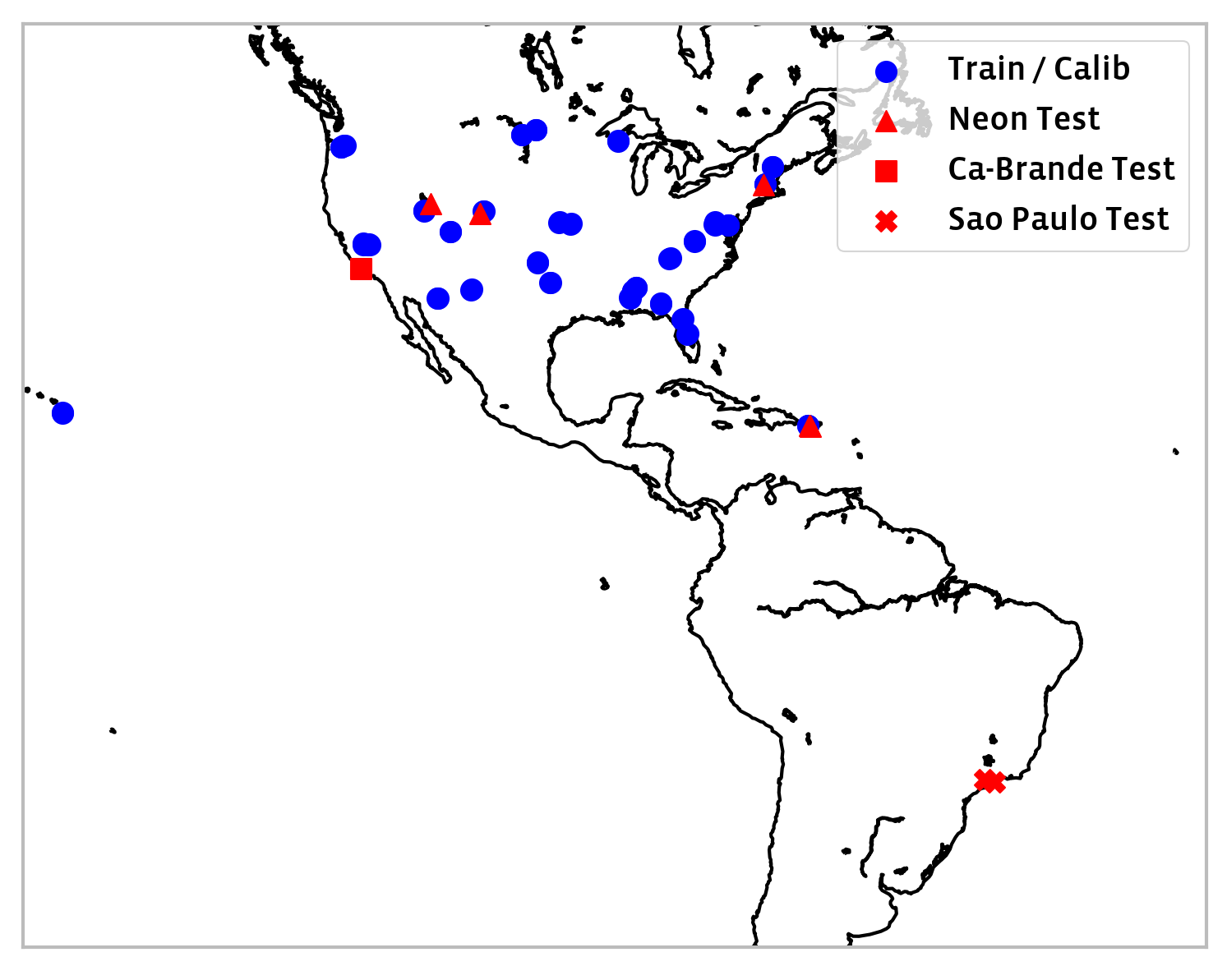}
       \caption{}
       \label{fig:train_test_val}
    \end{subfigure}%
    \begin{subfigure}[b]{0.5\textwidth}
        \centering
       \includegraphics[width=0.85\linewidth]{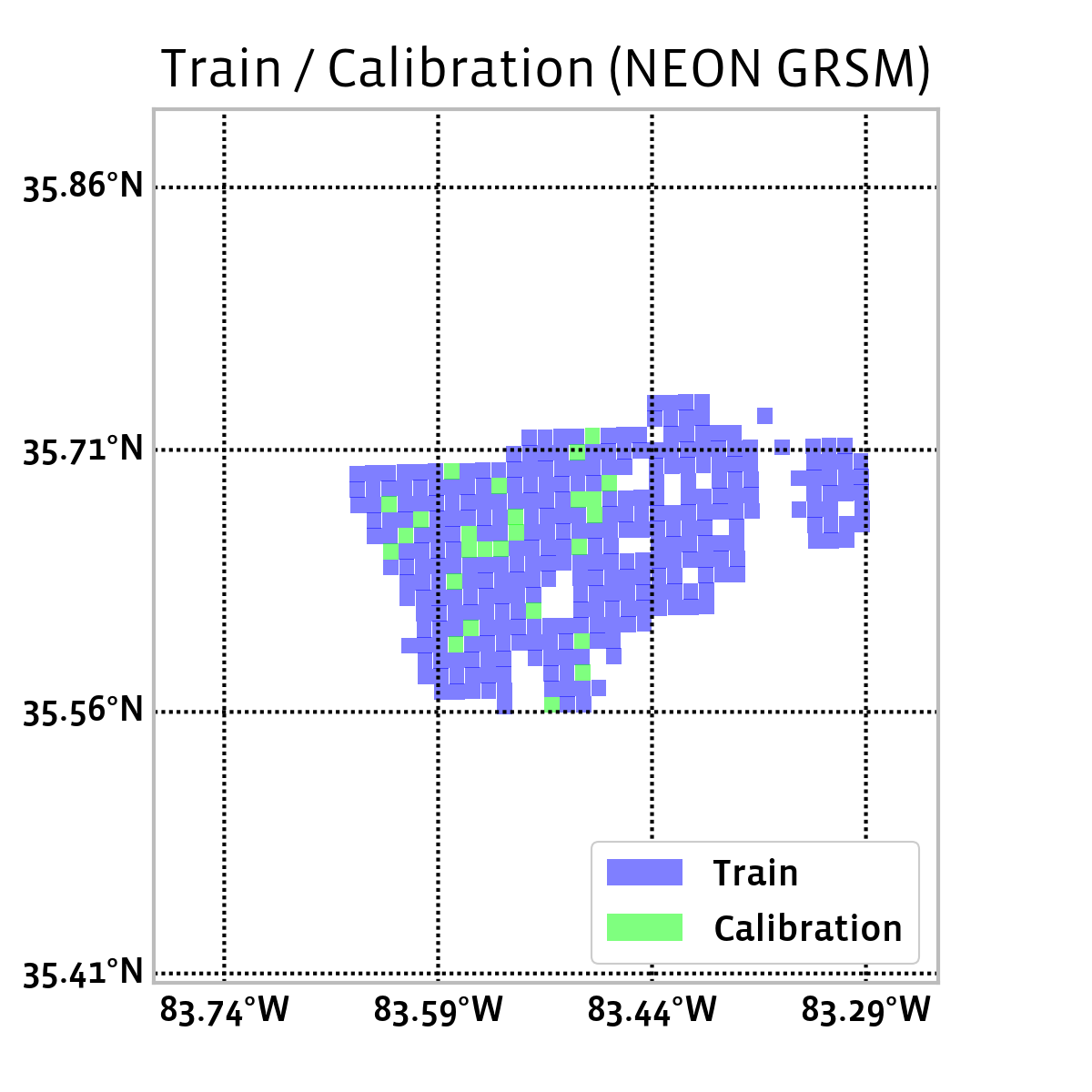}
       \caption{}
       \label{fig:train_val_GRSM}
    \end{subfigure}
\caption{Distribution of ALS Datasets: Train/Calibration/set-aside validation (aka Train/Validation/Test): (a) All ALS datasets. Here Train and Calibration points overlap and are shown in blue. Set-aside validation (Test) datasets are from non-overlapping geographic regions. (b) Zooming in on one Train / Calibration site (NEON GRSM) - we have randomly split the data into non spatially overlapping tiles so that calibration data is drawn from the same sites and ecosystems as training data, but separated spatially.}
\label{fig:dataset_distributions}
\end{figure}

The NEON sites during training  / calibration are : 
SJER, SOAP, TEAK, BART, DSNY, HARV, JERC, OSBS, DELA, GRSM, LENO, MLBS, BLAN, CLBJ, KONZ, NOGP, SCBI, TALL, UKFS, WOOD, ABBY, BONA, DEJU, JORN, MOAB, OAES, ONAQ, SERC, SRER, UNDE, WREF, HEAL, LAJA, RMNP, PUUM. 

The set-aside validation dataset, ``NEON test", contains the following NEON sites: CUPE, REDB, WLOU, HOPB, GUAN.

To ensure repeatability of our approach, we provide a complete list of CHM files used during training/calibration at:  \url{https://dataforgood-fb-data.s3.amazonaws.com/forests/v1/NEON_training_images.csv}

\section{GEDI Dataset and model training details}
\subsection{GEDI dataset}
\label{sec:gedi_data}

 The GEDI instrument is a full waveform lidar instrument aboard the International Space Station which has sampled global regions between 51.6$^{\circ}$ N $\&$ S latitude with a $\sim$25m beam footprint at ground surface. The instrument details are described in \cite{DUBAYAH2020100002}, and its measurements of canopy height are described in \cite{https://doi.org/10.3334/ornldaac/2056}.  We used the GEDI L2A Version 2 product and filtered the dataset to reduce noise by only including data which had: degrade flag $=0$, surface flag $=1$, solar elevation $<0$, and sensitivity $>0.95$. After this filtering, we were left with a total sample size of \num{1.3e9} measurements.  We used the $95^{\mbox{th}}$ percentile of relative height (RH95) that we paired to $128\times128$ pixel ($76\times76$ meter) satellite images from Maxar.  These images were processed as described in Section~\ref{sec:data_prep}, but were smaller to more closely approximate the GEDI footprint. Although these images are still significantly larger than the 25m GEDI footprint, we have found improved results from our GEDI model using larger areas - potentially due to pointing errors in the GEDI data and a larger spatial context improving the CNN model results. 

\subsection{Connection Between ALS CHM 95th percentiles and GEDI RH95}
\label{sec: connection between ALS P95 and GEDI RH95}

To leverage the GEDI model output, we made the following assumption: the GEDI model, on a $128\times128$ pixel sample, approximates the $95^{\mbox{th}}$ percentile (p95) of the sample's ground truth canopy height map. This is justified by running simulations with the the GEDI simulator from \cite{Hancock2019} on the NEON ALS point clouds. We used simulated values rather than actual GEDI measurements because the GEDI measurements suffer from point errors, and because the simlator allows for denser sampling from with the limited geographic footprint of our ALS dataset.

The GEDI RH95 measurement used for training the GEDI model corresponds to the $95^{\mbox{th}}$ percentile of the lidar's energy response. We simulated the GEDI RH95 values and find that they have excellent correlation ($R^2 = 0.88$) with the 95th percentile of the canopy height map around the corresponding GEDI footprints. This high correlation between GEDI RH95 and p95 of CHM was consistent across the diverse ecosystems covered in all 40 NEON sites in \ref{appendix:neonsites}.

\subsection{Height prediction network training}
\label{sec:gedi_training}
The GEDI measurements were split into a $80/10/10\%$ train/calibration/set-aside validation subsets. During training, the samples were drawn with a weight inversely proportional to the natural log of the total number of global samples in its RH95 bin, where each bin has a width of 1 meter. We found that this sampling method provided a good number of training sample from higher canopy height locations while not overly biasing the model towards ecosystems with the highest canopy heights. Log inverse sample weighting is a less aggressive re-weighting that typical linear inverse weighting, as done in \cite{Lang2022High}, which we choose so as not to overly bias the model towards the relatively few high canopy height samples.

After the  convolutional layers, we also input a collection of scalar values, designated as ``Satellite Metadata" in Figure~\ref{fig:train}. This metadata included: the latitude, longitude position of each image, the off-nadir view angle of the satellite, the angle between zenith and sun position at capture, and the terrain slope \citep{terraintiles} of the image footprint. Measured terrain slope is used during training, but set to zero height during forward inference, which allows the model to reduce the systematic error resulting from the bias of GEDI measurements towards higher canopy heights when the beam width straddles large surface gradients (see Section \ref{sec:gedi_comp}, \ref{sec:gedi_bias}).

When training the GEDI model, we only used random 90 degree rotations and random horizontal and vertical flips, since the larger volume of data made augmentation less helpful.

\begin{figure}[htb]
  \centering 
  \includegraphics[width=0.6\textwidth]{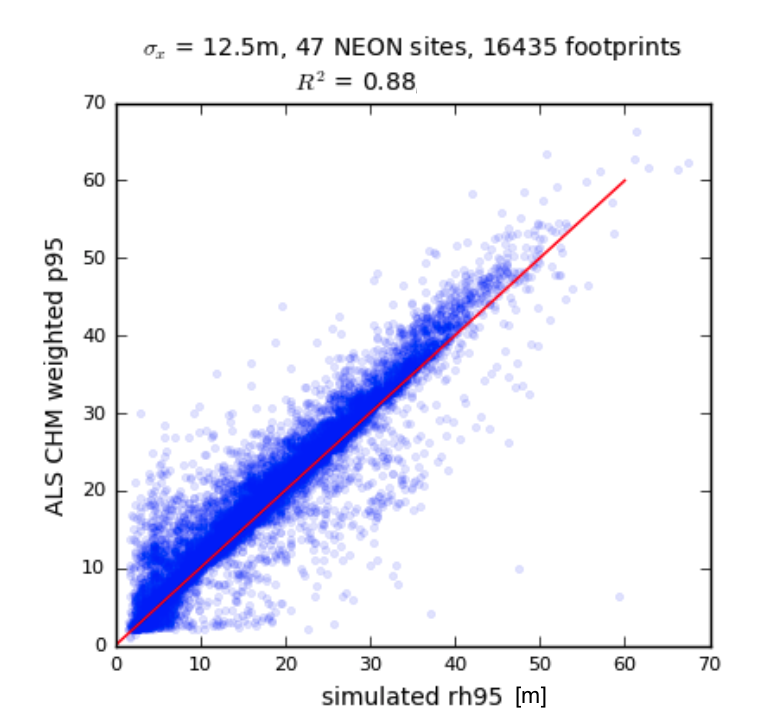}
  \caption{Correlation between 95th percentiles of ALS Canopy Height Maps and simulated GEDI RH95 values from the same maps. The 95th percentile is computed within weighted Gaussians with $\sigma=12.5m$, in order to roughly approximate the GEDI beam width.}
  \label{fig:als_rh95_corr}
\end{figure}

\subsection{GEDI height and terrain slope correlation}

\label{sec:gedi_bias}
As has been noted in \cite{rs12233948}, the GEDI instruments estimate of canopy height is influenced by the terrain slope. We found evidence of this correlation in the data, and due to this have chosen to set the terrain slope to zero during inference to mitigate this systematic.

\begin{figure}[ht]
  \centering 
  \includegraphics[width=.4\textwidth]{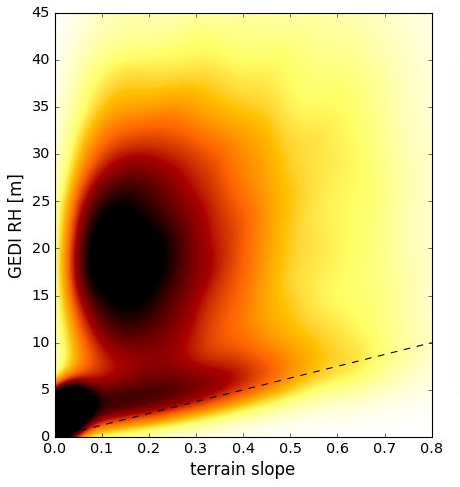}
  \caption{Correlation between terrain slope and GEDI RH95 for samples in CA. The dashed line indicates the height of the terrain with the GEDI beam (GEDI beam radius times the terrain slope). The heatmap is predominately above this line, indicating that there are no GEDI height estimates which fall below the terrain change within the beam.}
  \label{fig:gedi_bias}
\end{figure}

\section{Details on different metrics}
\label{appendix:metrics}

\subsection{Block $R^2$}
To compute the block $R^2$ score, we split the ground truth CHM $c$ and the prediction $\hat{c}$ into 50$\times$50 pixels blocks and average their values, leading to a 5$\times$5 array, reshaped into 1$\times$25 vectors $c^{(b)}$ and $\hat{c}^{(b)}$. Given the average ground truth CHM average of all $c^{(b)}$ in the test set, the classical $R^2$ score is then computed: \begin{equation}
R^2 _{\mbox{block}} = 1 - \frac{\sum(c^{(b)}_i - \hat{c}^{(b)}_i)^2}{\sum(c^{(b)}_i - \bar{c}^{(b)})^2}. 
\end{equation}

\subsection{Mean Error (ME)}

We compute the mean error, also referred as bias, as
\begin{equation}
\mbox{ME} = \frac{1}{|\mathcal{D}|} \sum_{i = 1 \dots |\mathcal{D}|} \hat{c}_i - c_i,
\end{equation}
where  $|\mathcal{D}|$ the number of pixels in the test set.

\subsection{Edge Error Metric (EE)}
\label{appendix:edgeacuracy}

\begin{figure}[htb]
  \centering 
  \includegraphics[width=1.0\textwidth]{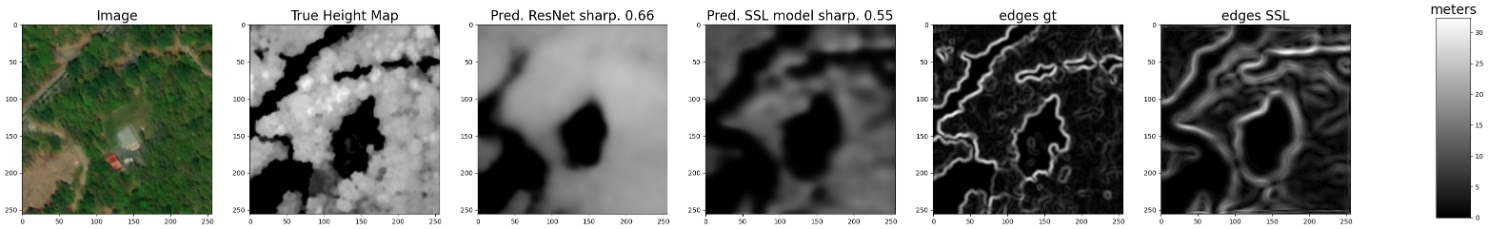}
  \caption{Illustration of edge error metric for two results: the ResUNet (trained with an L1 loss) edge error score is 0.66 in this example, the score of the SSL model is 0.55, computed using difference of the prediction and ground truth edge maps appearing in the two images at the right.}
  \label{fig:EE}
\end{figure}

We are interested in measuring the sharpness of the CHM while beeing close to the ground truth. Because a blurry prediction would lead to the same MAE, Block $R^2$ or PSNR than a sharp one, this metrics is not serving this purpose. Therefore, we established a metric comparing the image gradients of the maps, dubbed ``edge error score'', given by Algorithm~\ref{alg:ee}. Figure \ref{fig:EE} illustrates how this metric is computed in an example. 

\begin{algorithm}
\caption{Edge Error metric}\label{alg:ee}
\begin{algorithmic}
\State \textbf{1. Edge detection}
\Statex $E(\hat{c})$: Sobel detector on predicted CHM maps $\hat{c}$.
\Statex $E(c)$ : Sobel detector on GT CHM maps $c$.
\State \textbf{2. Compute normalization factor} 
$d = (\sum_i |E(\hat{c_i})|)+(\sum_i |E(c_i)|)$.
\State \textbf{3. Edge error score}
\Statex {\bf If} {$d>0$}
\Statex ~~~ score = $\frac{1}{d}  \sum_i |E(\hat{c_i})- E(c_i)|.$ 
\Statex {\bf Else}
\Statex ~~~ score = 0.
\end{algorithmic}
\end{algorithm}

\section{Architecture and Training Details}

\label{App:training_details}
Our code uses Pytorch 1.9.0 with Cuda 10.2. 

\paragraph{SSL pretraining}
We refer the reader \cite{Oquab2023SSL} for the SSL pretraining phase details. We only changed the image normalization parameters from ImageNet parameters to match the standard deviation and mean color intensities of our Satellite image dataset. The unsupervised pretraining of a large model took a little less than three days on two 8-GPUs Voltas. Instead of the standard ImageNet normalization parameters, we computed the mean and standard deviation on the dataset of 3.5M images. The large encoder contains 303 million parameters, while the huge one has 606 million.

\paragraph{Decoder training}
The training of CHM prediction from SSL features takes 8 hours for a large model on 8 GPUs, and 9 hours for a huge model. During this step, we kept the weights of the SSL encoder frozen and only train the DPT model. Our DPT decoder for the SSL model was trained for 140k steps using a Cosine learning rate schedule (from \num{1e-8} to \num{1e-4}) with a linear warmup step for 12k iterations. We used a batch size of 16. The decoder model contains 34.2 M of parameters. 

\paragraph{Estimating the carbon footprint of model training}

We estimate the carbon footprint of training the ViT huge model using the calculations from \cite{Oquab2023SSL}, a Thermal Design Power (TDP) of the V100-32G GPU equal to 250W, a Power Usage Effectiveness (PUE) of 1.1, a carbon intensity factor of 0.385 kg CO$_2$ per KWh, a time of 11 days $\times$ 24 hours $\times$ 64 GPUs = 16896 GPU hours. The 4647 kWh used to train the model is approximately equivalent to a CO$_2$ footprint of 4647 $\times$ 0.385 = 1.8T of CO$_2$. The training of the ResUNet baseline took 19 hours on 8 V100-32G GPUs, or approximately 16.1 kg of CO$_2$. The training of the decoder model took 75 GPU hours, generating about 8 kg of CO$_2$. While the carbon footprint of the ViT huge model, 1.81T of CO$_2$, was two orders of magnitude larger than the training of a ResUNet, the model training is a one-shot expense, and the inference time (and thus energy use and emissions) of the ViT huge and ResUNet were within the same order of magnitude. 

\paragraph{Architecture details}

Our encoder architecture is a ViT architecture as introduced by \cite{dosovitskiy2021image}. It treats an image as a set of patches, called tokens, that are first embedded into a feature space and then processed by a cascade of transformer layers to produce updated representations of the tokens. The transformer layers use multi-head attention and self attention as their fundamental operation. Multihead attention is an operation that relates each token to every token in the image and consequently, has a global receptive field. The ViT does not use down sampling operations in its intermediate stages and thus supports fine-grained feature maps also in the deeper layers of the network. 
For the huge model, the features consists in outputs from layers (9, 16, 22, 29). At each layer, a $8\times 8\times 1280$ feature map and $1\times 1\times 1280$ class output is extracted.
In the DPT decoder, the set of tokens at various stages of the backbone is first reassembled into image like representations. Then, the decoder iteratively fuses the feature maps from different stages and produces the final dense prediction using an application specific output head. This step involves several residual convolution layers. The code of our backbone is available at \url{https://github.com/facebookresearch/dinov2}, with pre-training on natural images. 

\section{Alternate Loss Function Ablation}
\label{appendix:loss}

We compare in Table \ref{tab:ablation_all} results of models trained with L1 loss or Sigloss, and using different sizes of pretraining dataset: 
one with \num{3.5e6} images (referred to as ``3.5M") and one with \num{18e6} images (``18M"). More pretraining data improves the performance of the SSL models. In terms of loss, we did not notice strong difference between L2 and sigloss, while the L1 results were slightly worse.

\begin{table}[ht]
\resizebox{\columnwidth}{!}{
\begin{tabular}{ccccccccccccccccc}
\toprule
& \multicolumn{4}{c}{\bf{Neon test}} & \multicolumn{4}{c}{{\bf S\~{a}o Paulo}}  &  \multicolumn{4}{c}{{\bf Average}}\\
& MAE & $R^2$ & ME & EE& MAE & $R^2$ &  ME &EE&  MAE & $R^2$  & ME & EE\\
\midrule 
3.5M sl & 2.8 & 0.67 & -1.2 & 0.51 & 6.0 & 0.14 & -4.2 & 0.60 & 3.1 &0.56&1.9 & 0.54\\
18M sl & 2.9 & 0.66 & -1.3 & 0.52 & 4.9 & 0.46 & -2.1 & 0.59 &2.9 & 0.64& 1.3 & 0.54 \\
\midrule
18M linear sl & 3.0 & 0.58 & -1.8 & 0.68 & 7.1 & -0.27 & -6.7 & 0.71 & 3.6 & 0.41 & 2.8 & 0.67\\
\midrule 
18M cl sl & 2.6 &	0.71	&-0.9 &{\bf	0.48}	&{\bf4.9}&	{\bf0.47}&	-1.9&	{\bf0.55}	&{\bf2.7	}&0.70&	1.0&	{\bf 0.51}\\
18M cl l1 & {\bf 2.5}	& 0.80 &	{\bf0.0}&	0.51&	5.2&	0.39&	-2.6& 
0.56 & 2.9	& 0.72	&0.7	& 0.53\\
18M cl l2 & 2.6	&{\bf0.86}	&-0.1&	0.52	&5.1	&0.43&	{\bf-1.4}&	{\bf0.55}&2.8&	{\bf0.75}&	{\bf0.5}&	{\bf 0.51}\\
\bottomrule 
\end{tabular}}
\caption{CHM prediction accuracy metrics with different loss functions. sl: sigloss. cl : using classification output. linear: using a linear layer instead of DPT. We do not display CA Brande result to improve visibility but the results are included in the average.}
\label{tab:ablation_all}
\end{table}

\section{Residuals with respect to the GEDI RH95}
\label{app:gedi_plot}

\begin{figure}[htb]
  \centering 
\includegraphics[width=1\linewidth]{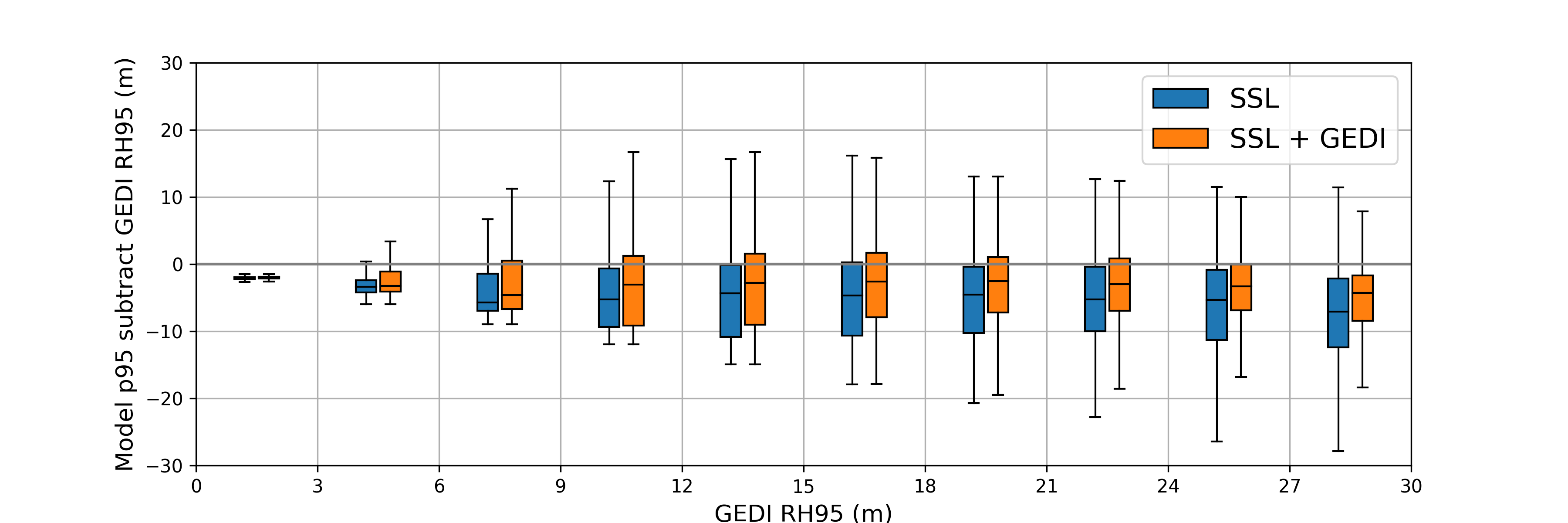}
 \caption{Residuals of p95 CHM predictions with GEDI RH95, with respect to the GEDI RH95.}
  \label{fig:gedirh95}
\end{figure}

Figure \ref{fig:gedirh95} displays the difference between the p95 of block model CHM predictions and the measured GEDI RH95 metrics w.r.t the GEDI RH95.

\section{Normalization for inference on aerial imagery}
\label{appendix:normalization}

An image normalization step is necessary to improve the SSL inference performance on aerial images, when training only on Satellite imagery. We perform a classical histogram normalization of the aerial images (i.e. normalize the RGB channels of the aerial image to the p5-p95 distribution of the satellite image). This makes the color balance much more similar, leading to better performance for the SSL model. The satellite image is taken through much more atmosphere and we expect it to be less blue on average, because of preferential scattering of shorter wavelengths.
Noting $I$ the satellite image, $A$ the original aerial image, we first compute for each color channel $i$ and each image $X$
the $5\%$ percentile $p_5(X)_i$ and  $95\%$ percentile $p_{95}(X)_i$. Then, the normalized aerial image is given by
\[
A_i = (A_i-p_5(A)_i) * \frac{p_{95}(I)_i-p_{5}(I)_i}{p_{95}(A)_i-p_{5}(A)_i} + p_5(I)_i.
\]
We only apply this normalization to the SSL model trained on satellite imagery. Applying this normalization to the ResUNet model deteriorated the results. 

\Urlmuskip=0mu plus 1mu\relax

{
\small{

}
}

\end{document}